\documentclass[10pt,twocolumn,letterpaper]{article}

\usepackage[pagenumbers]{cvpr} 

\usepackage[dvipsnames]{xcolor}

\usepackage[hang,flushmargin]{footmisc}

\definecolor{cvprblue}{rgb}{0.21,0.49,0.74}
\usepackage[pagebackref,breaklinks,colorlinks,citecolor=cvprblue]{hyperref}

\def\paperID{94} 
\def\confName{CVPR}
\def\confYear{2024}

\definecolor{color_green}{HTML}{92D050}

\newcommand\blfootnote[1]{%
    \begingroup 
    \renewcommand\thefootnote{}\footnote{#1}%
    \addtocounter{footnote}{-1}%
    \endgroup 
}

\begin{document}
\title{Intelligent Grimm - Open-ended Visual Storytelling via Latent Diffusion Models}

\author{Chang Liu$^{1,3*}$, Haoning Wu$^{1*}$, Yujie Zhong$^{2}$, Xiaoyun Zhang$^{1}$, Yanfeng Wang$^{1,3}$, Weidi Xie$^{1,3}$\\[3pt]
$^{1}$Coop. Medianet Innovation Center, Shanghai Jiao Tong University, China\\[2pt]
$^{2}$Meituan Inc., China \hspace{0.5cm} $^{3}$Shanghai AI Laboratory, China\\[2pt]
}

\twocolumn[{%
\renewcommand\twocolumn[1][]{#1}%
\maketitle
\vspace{-1.0cm}
\begin{center}
   \centering
   \includegraphics[width=\textwidth]{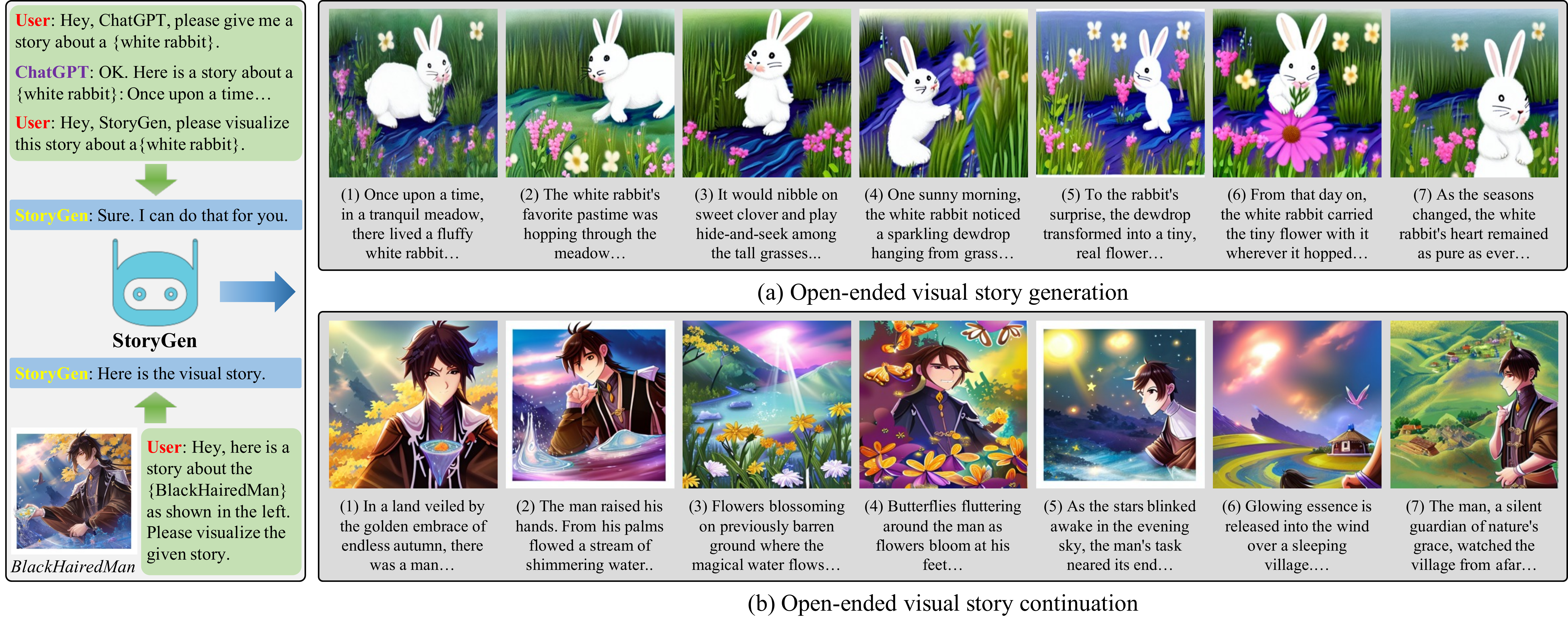} 
   \vspace{-0.7cm}
   \captionof{figure}{
   \textbf{An illustration of open-ended visual storytelling.}
   In practice, users can feed a unique and engaging story synthesized by a large language model into our proposed {\bf StoryGen} model to generate a sequence of images coherently, denoted as {\em open-ended visual story generation}. 
   And they can also provide a pre-defined character with its corresponding storyline, to perform {\em open-ended visual story continuation}.
   We recommend the reader to zoom in and read the story. 
   }
  \label{fig:teaser}
 \end{center}
 }]

 \blfootnote{*: These authors contribute equally to this work. \\ 
}

\begin{abstract}
Generative models have recently exhibited exceptional capabilities in text-to-image generation, but still struggle to generate image sequences coherently. 
In this work, we focus on a novel, yet challenging task of generating a coherent image sequence based on a given storyline, denoted as \textbf{open-ended visual storytelling}.
We make the following three contributions:
(i) to fulfill the task of visual storytelling,
we propose a learning-based auto-regressive image generation model, termed as \textbf{StoryGen}, with a novel vision-language context module, that enables to generate the current frame by conditioning on the corresponding text prompt and preceding image-caption pairs;
(ii) to address the data shortage of visual storytelling, 
we collect paired image-text sequences by sourcing from online videos and open-source E-books, establishing processing pipeline for constructing a large-scale dataset with diverse characters, storylines, and artistic styles, named \textbf{StorySalon};
(iii) Quantitative experiments and human evaluations have validated the superiority of our StoryGen, where we show StoryGen can generalize to unseen characters without any optimization, and generate image sequences with coherent content and consistent character.
Code, dataset, and models are available at \url{https://haoningwu3639.github.io/StoryGen_Webpage/}.
\end{abstract}


\noindent {\em  ``Mirror mirror on the wall, who's the fairest of them all?" }

\vspace{3pt}{\em \hspace{4.3cm}------ Grimms' Fairy Tales}
\section{Introduction}

This paper considers an interesting, yet challenging task, namely, {\em open-ended visual storytelling}. 
The goal is to train a generative model that effectively captures the relation between visual elements and corresponding text descriptions, to generate a sequence of images that tell a visually coherent story, as shown in Figure~\ref{fig:teaser}. 
The outcome of this task has significant potential for education, 
as it provides children with an engaging and interactive way to learn complex visual concepts and develop imagination, creativity, emotional intelligence, and language skills, as evidenced by research in psychology~\cite{david2012how, strouse2018role}.

The recent literature has witnessed tremendous progress in image generation, particularly with the guidance of text as prompt, such as stable diffusion~\cite{SDM}, DALL·E~\cite{ramesh2021zero} and Imagen~\cite{ho2022imagen}.
However, to generalize the models for open-ended visual storytelling, 
we are facing three challenges:
(i) previous models are designed to only generate images independently, 
without considering context, for example, preceding frames or overall narrative, 
resulting in a lack of visual consistency;
(ii) most methods generate images by only conditioning on text, 
which potentially leads to ambiguities or requires unnecessarily long descriptions to maintain character appearances;
(iii) existing datasets are limited to a few animations, covering a closed set of vocabulary or characters~\cite{li2019storygan,maharana2022storydall,pan2022synthesizing}.
Training on such datasets suffers from severe overfitting on seen characters,
leading to unsatisfactory generalization capability for open-ended generation.


This paper describes a learning-based model for open-ended visual storytelling, termed as \textbf{StoryGen}, 
that enables to generate unseen characters without any further optimization, while having character consistency. 
At inference, StoryGen can synthesize frames either by taking text prompts, or along with preceding image-text pairs as conditions, {\em i.e.}, iteratively creating visual sequences that are aligned with language description, while being consistent with preceding frames in both style and character perspectives.
Specifically, to achieve consistency within the generated image sequence,
we incorporate a novel \textbf{vision-language context module} into the pre-trained stable diffusion model, which provides visual context by conditioning the generation process on extracted diffusion denoising feature of previous frames under the guidance of corresponding captions. 


As for training, we construct a dataset called \textbf{StorySalon}, 
that features a rich source of coherent images and stories, primarily comprising children's storybooks collected from videos and E-books. 
As a result, our dataset includes a diverse vocabulary with different characters, storylines, and artistic styles. 
The scale and diversity of our collected dataset enable the model for open-vocabulary visual storytelling, 
{\em i.e.}, generating new image sequences that are not limited to pre-defined storylines, characters, or scenes. 
For example, we can prompt a large language model to create unique and engaging stories, then feed them into StoryGen for generation, as shown in Figure~\ref{fig:teaser}.

To summarize, we make the following contributions in this paper:
(i) we initiate a fun yet challenging task, namely, {\em open-ended visual storytelling}, that involves generating engaging image sequences aligned to a given storyline;
(ii) we propose a learning-based open-ended visual storytelling model, termed as \textbf{StoryGen}, which can generalize to unseen characters without any further optimization and generate coherent visual stories, utilizing a novel vision-language context module;
(iii) we establish a data processing pipeline and collect a large-scale dataset of storybooks, called \textbf{StorySalon}, from online videos and open-source E-books, resulting in a diverse vocabulary with various characters, storylines, and artistic styles;
(iv) we conduct quantitative experiments and human evaluations to validate the effectiveness of our proposed modules, demonstrating the superiority of our model, in terms of image quality, consistency, and visual-language alignment of generated contents.

\vspace{-3pt}
\section{Related Works}
\label{sec:formatting}
\noindent {\bf Text-to-image Generation} 
has been tackled using various generative models, with 
GAN~\cite{goodfellow2020generative} as the first widely used model. 
Several GAN-based methods~\cite{zhang2017stackgan, zhang2018stackgan++, xu2018attngan} have achieved notable success, and auto-regressive transformers~\cite{vaswani2017attention}, such as DALL·E~\cite{ramesh2021zero}, have also demonstrated the ability to generate high-quality images based on text prompts. 
Recently, diffusion models, such as Imagen~\cite{saharia2022photorealistic} and DALL·E 2~\cite{ramesh2022hierarchical}, have emerged as a popular approach.
Stable Diffusion Models~\cite{SDM} performs diffusion process in latent space, and can generate impressive images after pre-training on a large-scale text-image dataset.

\vspace{3pt} 
\noindent {\bf Diffusion Models} learn to model a data distribution via iterative denoising and are trained with denoising score matching.
Notably, DDPM~\cite{ho2020denoising} demonstrates improved performance over other generative models, while DDIM~\cite{song2020denoising} significantly boosts efficiency.
In view of their superior generative capabilities, diffusion models have found extensive utility in various downstream applications besides image generation, such as video generation~\cite{esser2023structure,ho2022video,ho2022imagen,singer2022make}, image manipulation~\cite{brooks2022instructpix2pix,meng2021sdedit,kawar2022imagic,hertz2022prompt}, grounded generation~\cite{li2023guiding}, image restoration~\cite{chen2023hierarchical}, and image inpainting~\cite{xie2022smartbrush,nichol2021glide,lugmayr2022repaint,avrahami2022blended}.
\begin{figure*}[t]
  \centering
  \includegraphics[width=\textwidth]{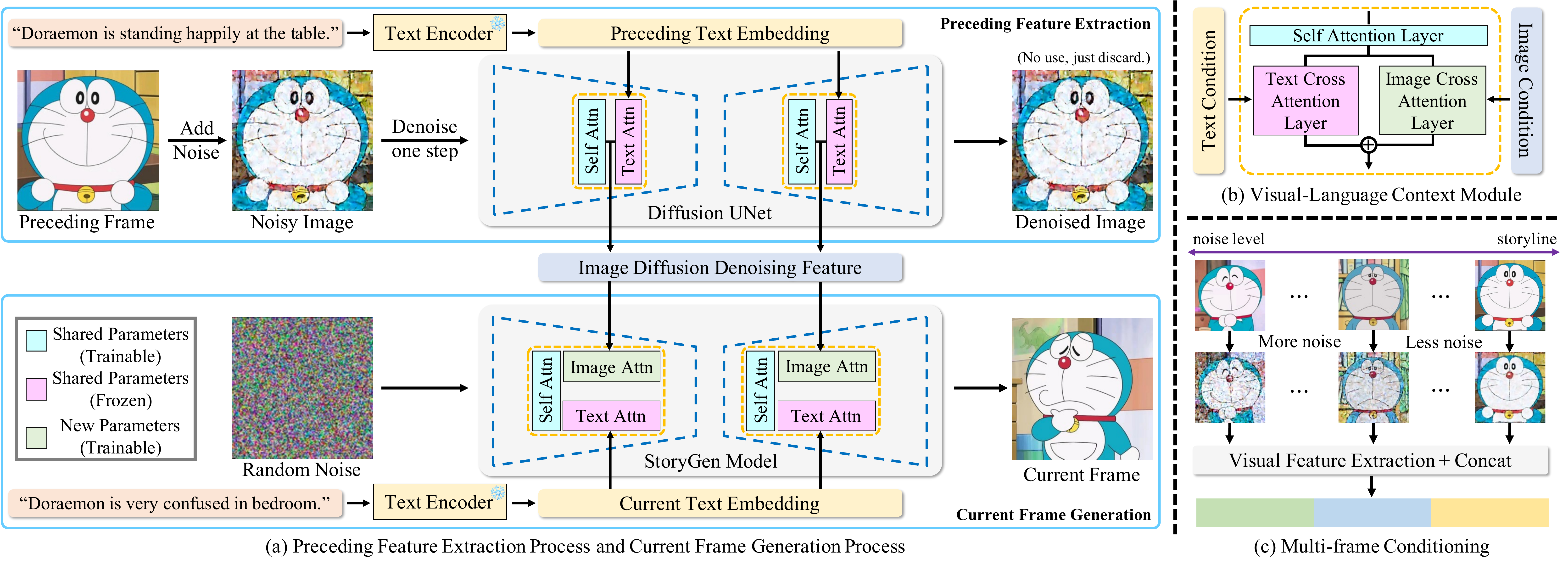} \\
  \vspace{-4pt}
  \caption{
  \textbf{Architecture Overview}. 
  (a) Our StoryGen model utilizes current text prompt and previous visual-language contexts as conditions to generate an image, iteratively synthesizing a coherent image sequence.
  Note the parameters of the corresponding attention layers are shared between Diffusion UNet and StoryGen.
  To avoid potential ambiguity, the parameters are not shared across UNet blocks in a single model.
  (b) The proposed Visual-Language Context Module can effectively combine the information from current text prompt and contexts from preceding image-caption pairs.
  (c) We add more noise to reference frames with longer temporal distances to the current frame as positional encoding to distinguish the temporal order. 
  The multiple features can then be directly concatenated to serve as context conditions.
    }
    \vspace{-6pt}
 \label{fig:arch}
\end{figure*}

\vspace{2pt} 
\noindent {\bf Story Synthesis} is first introduced as the task of story visualization by StoryGAN~\cite{li2019storygan},
which presents a GAN-based framework and the PororoSV dataset, derived from cartoons. 
Some works~\cite{maharana2021improving, maharana2021integrating} follow the GAN-based framework, whereas others~\cite{li2022word, chen2022character} emphasize more on text representation.
StoryDALL-E~\cite{maharana2022storydall} extends story synthesis to story continuation with the initial image given, and exploits a pre-trained DALL·E model~\cite{ramesh2021zero} to produce coherent images.
AR-LDM~\cite{pan2022synthesizing} introduces an auto-regressive latent diffusion model to generate image sequences, but only consistent within a limited character vocabulary.
NUWA-XL~\cite{yin2023nuwa} exploits hierarchical diffusion models to synthesize long videos, but still achieve character consistency by memorizing.
TaleCrafter~\cite{gong2023talecrafter} proposes a story visualization system and utilizes LoRA~\cite{hu2021lora} to achieve character consistency. However, large-scale applications will be constrained due to its optimization-based nature.
In this paper, we target more ambitious applications, to develop an open-ended visual storytelling model,
that can synthesize coherent image sequences based on storylines of diverse topics.

\section{Method}
In this section, we start by formulating the problem of open-ended visual storytelling in Section~\ref{sec:problem};
then we elaborate on the proposed StoryGen architecture in Section~\ref{sec:arch};
lastly, we present details for model training in Section~\ref{sec:training}.

\subsection{Problem Formulation}
\label{sec:problem}
In this paper, we focus on a challenging task, termed as {\em open-ended visual storytelling}, the goal is to generate continuous image sequence from a given story in the form of natural language.
Specifically, we propose a learning-based auto-regressive image generation model, called \textbf{StoryGen}, that generates the current frame $\hat{\mathcal{I}}_k$ by conditioning on the current text prompt $\mathcal{T}_k$, and image-text pairs $(\hat{\mathcal{I}}_{<k},\mathcal{T}_{<k})$ of previous frames,
as illustrated in Figure~\ref{fig:arch} (a).
The model is formulated as follows: 
\vspace{-4pt}
\begin{align*}
\{\hat{\mathcal{I}}_1, \hat{\mathcal{I}}_2, \dots, \hat{\mathcal{I}}_L\} 
&= \Phi_{\text{StoryGen}}(\{\mathcal{T}_1, \mathcal{T}_2, \dots, \mathcal{T}_L\}; \Theta) \\
\hat{\mathcal{I}}_k & := \Phi_{\text{StoryGen}} (\hat{\mathcal{I}}_{k}| \mathcal{T}_k, (\hat{\mathcal{I}}_{<k}, \mathcal{T}_{<k}))
\end{align*}
Here, $\{\mathcal{T}_1, \mathcal{T}_2, \dots, \mathcal{T}_L\}$ refer to the given storylines, and $\{\hat{\mathcal{I}}_1, \hat{\mathcal{I}}_2, \dots, \hat{\mathcal{I}}_L\}$ denote the generated image sequence.
$\Phi_{\text{StoryGen}}(\cdot)$ represents our proposed StoryGen model.
In one-step generation, StoryGen takes the current text prompt, 
and preceding image-caption pairs as conditions, and generates an image consistent with both the story's narrative and previous frames. The whole image sequence can then be synthesized with step-by-step inference.

\vspace{2pt} \noindent \textbf{Relation to Existing Tasks.}
In contrast to existing story visualization works, 
this paper makes improvements from two aspects:
(i) conventional generation/continuation tasks are limited to training on specific characters/stories, for example, 
~\cite{li2019storygan,maharana2022storydall,pan2022synthesizing} only exploits datasets from animation {\em The Flintstones} and {\em Pororo}, while our model enables to generate visual stories based on any given storyline, such as a brand-new one generated by ChatGPT; and any pre-defined character, for example, {\em `Doraemon'} from the Internet; (ii) unlike existing work that requires costly character-specific optimization, for example, \cite{pan2022synthesizing,gong2023talecrafter} rely on LoRA-based~\cite{hu2021lora} optimization to adapt to new characters, our model is learning-based and expected to generalize to any unseen character without any further optimization. 

\subsection{Architecture}
\label{sec:arch}

To tackle the problem of open-ended visual storytelling, 
we expect the model to not only condition on the current text prompt, 
but also preceding image-text pairs.
In this section, we describe the procedure for one-step generation,
{\em i.e.}, generating the $k$-th frame~($k>1$) by conditioning on
$\{ (\hat{\mathcal{I}}_1, \mathcal{T}_1), \dots, (\hat{\mathcal{I}}_{k-1}, \mathcal{T}_{k-1}), \mathcal{T}_k\}$. Generally speaking, our proposed \textbf{StoryGen} model comprises four components: 
(i) Input Initialization,
(ii) Context Encoding, 
(iii) Visual-Language Contextual Fusion, 
(iv) Conditional Generation. 

\vspace{2pt} \noindent \textbf{Input Initialization.}
Our model is built upon the foundation of a pre-trained stable diffusion model~(SDM), which randomly samples a noisy latent $\mathbf{x}$ from the latent space of the VAE~\cite{kingma2013auto} encoder. Moreover, for a given text prompt $\mathcal{T}_k$, 
the text condition will be extracted by a pre-trained CLIP~\cite{CLIP} text encoder $\phi_{\text{CLIP}}$ via 
$\mathcal{C}^{\text{T}} = \phi_{\text{CLIP}}(\mathcal{T}_k)$.

\vspace{2pt} 
\noindent \textbf{Context Encoding.}
In standard SDM, the noisy latent is recursively denoised with a UNet, conditioning on the text prompt. However, in our case, it is crucial for the generation procedure to also condition on context features of preceding frames, to maintain consistency in characters and storyline.

In practice, to extract the contextual features, we add noise to the preceding frames and exploit the pre-trained SDM to denoise for one diffusion step under the guidance of their corresponding captions. The diffusion features after every self-attention layer in the UNet blocks can be directly selected to serve as the conditioning visual context features, thus constituting a pyramid of visual context features.
The visual condition features for $\hat{\mathcal{I}}_k$ can be expressed as: 
\begin{align*}
    \mathcal{C}^{\text{V}} = [\phi_{\text{SDM}}(\hat{\mathcal{I}}_{1}, \phi_{\text{CLIP}}(\mathcal{T}_{1})), \dots, \phi_{\text{SDM}}(\hat{\mathcal{I}}_{k-1}, \phi_{\text{CLIP}}(\mathcal{T}_{k-1}))]
\end{align*}

Experimentally, we notice that, the magnitude of noise added to the preceding frames can greatly affect the conditional generation quality,
{\em i.e.}, large-scale noise on preceding frames incurs severe information loss. 
Thus, we propose to use a much smaller diffusion timestep $t'$ for preceding frames compared with the diffusion timestep $t$ of the current image $\hat{\mathcal{I}}_{k}$, and follow a $t' = t / 10$ rule.
As depicted in Figure~\ref{fig:arch} (c), 
in case of multiple preceding image-caption pairs, 
we use larger $t'$ for frames with longer temporal distances to $\hat{\mathcal{I}}_{k}$. Therefore, the extracted multi-frame visual context features can be directly concatenated, and their different noise level will serve as temporal positional embedding.
Such design reflects the intuition that frames with longer distances will incur less effect on generating the current frame.

\vspace{2pt} \noindent \textbf{Vision-Language Contextual Fusion.}
Here, our vision-language context module is designed to fuse information from current text prompt and contextual information from preceding image-caption pairs.
This is achieved by augmenting the transformer decoder in SDM with an additional image cross-attention layer. 
Note that, the math expression in this section is not strict, we omit the footnote of diffusion timestep $t$ and UNet block level $l$ for simplicity.


Specifically, on visual context conditioning, 
the noisy latent $\mathbf{x}$ is projected into query, and cross-attends to the visual context features from the corresponding-level UNet block that act as key and value, 
denoted as: 
\begin{equation*}
   \mathbf{Q}_I=\mathbf{x}\mathbf{W}_I^Q, \quad \mathbf{K}_I=\mathcal{C}^{\text{V}}\mathbf{W}_I^K, \quad \mathbf{V}_I=\mathcal{C}^{\text{V}}\mathbf{W}_I^V
\end{equation*}
where $\mathbf{W}_I^Q$, $\mathbf{W}_I^K$, and $\mathbf{W}_I^V$ represent different projection matrices, respectively.

On text conditioning, the noisy latent $\mathbf{x}$ is again projected to query, 
and cross-attends to the text features of the current prompt encoded by CLIP text encoder, {\em i.e.},
\begin{equation*}
   \mathbf{Q}_T=\mathbf{x}\mathbf{W}_T^Q, \quad \mathbf{K}_T=\mathcal{C}^{\text{T}}\mathbf{W}_T^K, \quad \mathbf{V}_T=\mathcal{C}^{\text{T}}\mathbf{W}_T^V
\end{equation*}
where $\mathbf{W}_T^Q$, $\mathbf{W}_T^K$, and $ \mathbf{W}_T^V$ also represent corresponding projection matrices.


As depicted in Figure~\ref{fig:arch} (b), the image cross-attention layer is inserted in parallel to the text cross-attention layer in the transformer decoder of UNet blocks. Drawing inspiration from ControlNet~\cite{controlnet}, the results from these two cross-attention layers are simply summed up as the final output $\mathbf{O}$.
The final output can thus be expressed as:
\begin{equation*}
    \mathbf{O} = \text{Softmax}(\frac{\mathbf{Q}_I(\mathbf{K}_I)^\top}{\sqrt{d}})\mathbf{V}_I + \text{Softmax}(\frac{\mathbf{Q}_T(\mathbf{K}_T)^\top}{\sqrt{d}})\mathbf{V}_T
\end{equation*}


\noindent \textbf{Conditional Generation.}
With the fused vision-language condition features from above, 
our StoryGen can now generate visual stories that achieve both content coherence and character consistency.
Here, our conditional generation procedure can be represented as: 
\begin{align*}
    \hat{\mathcal{I}}_k = \Phi_{\text{StoryGen}} (\hat{\mathcal{I}}_{k}| \mathcal{T}_k, (\hat{\mathcal{I}}_{<k}, \mathcal{T}_{<k}))
    = \Phi_{\text{StoryGen}}(\mathbf{x}, \mathcal{C}^{\text{T}}, \mathcal{C}^{\text{V}})
\end{align*}

With the new conditioning modality introduced, we also adopt another classifier-free guidance term~\cite{ho2022classifier}, 
as has been done in ~\cite{brooks2022instructpix2pix}.
Concretely, we exploit two different guidance scales, $w_v$ and $w_t$ for the visual condition and the text condition.
The relation between the final noise for inference $\bar{\boldsymbol{\epsilon}}_\theta$ and UNet-predicted noise $\boldsymbol{\epsilon}_\theta$ is now expressed as:
\begin{align*}
\bar{\boldsymbol{\epsilon}}_\theta(\mathbf{x}_t, t, \mathcal{C}^{\text{V}} , \mathcal{C}^{\text{T}})
 &  = \boldsymbol{\epsilon}_\theta(\mathbf{x}_t, t, \varnothing, \varnothing) \nonumber \\
 & + w_v(\boldsymbol{\epsilon}_\theta(\mathbf{x}_t, t, \mathcal{C}^{\text{V}} ,\varnothing) - \boldsymbol{\epsilon}_\theta(\mathbf{x}_t, t, \varnothing, \varnothing)) \nonumber \\
 & + w_t(\boldsymbol{\epsilon}_\theta(\mathbf{x}_t, t, \mathcal{C}^{\text{V}} ,\mathcal{C}^{\text{T}}) - \boldsymbol{\epsilon}_\theta(\mathbf{x}_t, t, \mathcal{C}^{\text{V}}, \varnothing))
\end{align*}

\noindent \textbf{Discussion.}
Our work differs from previous ones from two aspects.
First, our StoryGen is a learning-based method, 
which can directly generalize to unseen characters by attending to reference images.
Second, we propose to condition the generation process on diffusion features of preceding image-text pairs from the same SDM,
which preserves more visual details, greatly differing from existing works~\cite{ye2023ip-adapter,li2023blip,li2023videogen} using CLIP, BLIP~\cite{li2022blip}, or VAE features.



\subsection{Model Training}
\label{sec:training}
\noindent {\bf Training Objective.}
At training stage, we randomly sample a triplet each time, 
{\em i.e.}, $\{\mathcal{I}_k, \mathcal{T}_k, (\mathcal{I}_{<k}, \mathcal{T}_{<k})\}$.
The objective function can be expressed as:
\begin{align*}
\mathcal{L}_t = \mathbb{E}_{t \sim [1, T], \mathbf{x}_0, \boldsymbol{\epsilon}_t,\mathcal{C}^{\text{V}} , \mathcal{C}^{\text{T}}} \Big[\|\boldsymbol{\epsilon}_t - \boldsymbol{\epsilon}_\theta(\mathbf{x}_t, t,\mathcal{C}^{\text{V}}, \mathcal{C}^{\text{T}})\|^2 \Big]
\end{align*}

\noindent{\bf Two-stage Training Strategy.}
Our two-stage training strategy includes single-frame pre-training and multiple-frame fine-tuning. 
To be specific, at the first stage, we do not introduce additional image cross-attention layers, and only train self-attention layers in standard SDM to ensure the single-frame generation ability.
In multiple-frame fine-tuning, we train additional image cross-attention layers in vision-language context module on our dataset, with all other parameters frozen. This enables the generation procedure to utilize information from not only current prompt, 
but also preceding image-caption pairs.





\vspace{3pt}
\noindent{\bf Inference.} 
As shown in Figure~\ref{fig:teaser}, at inference time, 
we can prompt ChatGPT to generate novel storylines, and synthesize the first image directly
or attending to a pre-defined character.
Then the previously synthesized frames, along with the story descriptions, are treated as conditions to synthesize the image sequence in an auto-regressive manner. Experimentally, our proposed StoryGen is shown to generate images that align with the storyline, as well as maintain consistency with previously generated frames.

\begin{figure}[t]
  \centering
  \includegraphics[width=0.45\textwidth]{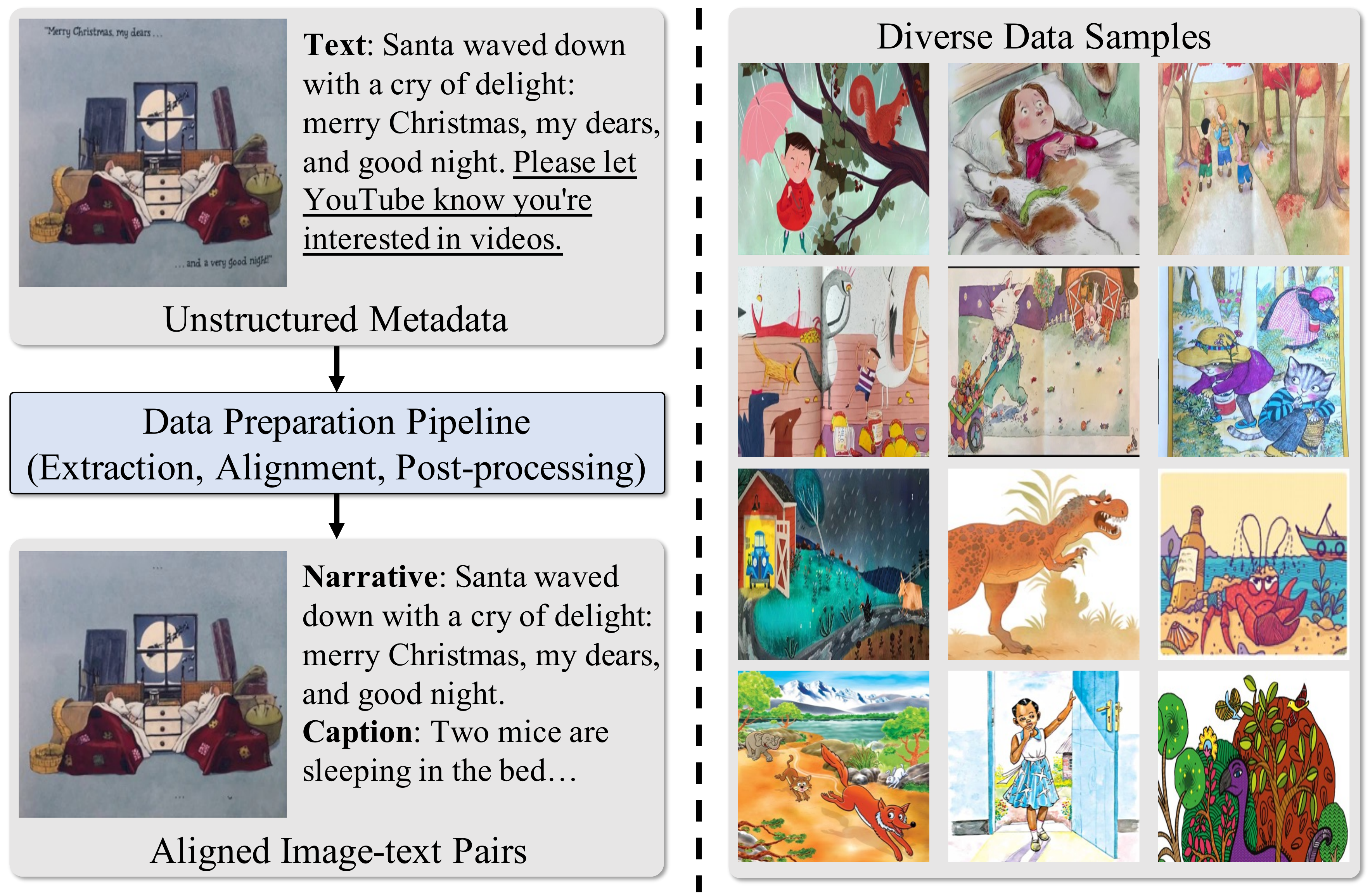} 
  \vspace{-4pt}
  \caption{
  \textbf{Dataset Pipeline and Visualization}.
  \textbf{Left}: Metadata sourced from the Internet undergoes a three-step pipeline including frame extraction, visual-language alignment and post-processing, resulting in properly aligned image-text pairs.
  \textbf{Right}: Our StorySalon dataset contains diverse styles and characters.
  }
 \label{fig:dataset}
 \vspace{-6pt}
\end{figure}

\section{StorySalon Dataset}
\label{sec:dataset}
In order to train our proposed {\em open-ended visual storytelling} model, 
we construct a large-scale dataset, termed as {\bf StorySalon}.
The dataset contains videos and E-books with diverse characters, storylines, and artistic styles.
Specifically, we download a large number of videos and subtitles from YouTube, by querying keywords related to story-telling for children, for instance, {\em storytime}.
Additionally, we collect E-books~(partially with corresponding audios available) from six open-source libraries which are all registered under the Creative Commons 4.0 International Attribution (CC BY 4.0) license. 
In the following, we elaborate on the data processing pipeline and statistics of our collected dataset. 

\vspace{2pt}\noindent{\bf Visual Frame Extraction.}
We extract keyframes from the videos, along with the corresponding subtitles and their timestamps. To remove duplicate frames, we extract ViT features for each frame using pre-trained DINO~\cite{Mathilde21}. 
For the image groups with high similarity scores, we only keep one of each. 
Then, we use YOLOv7~\cite{wang2022yolov7} to segment and remove real-person frames and headshots, as they often correspond to the story-teller and are unrelated to the content of the storybook. 
Similarly, we extract images from the downloaded E-books, 
except for those with extraneous information, 
for example, the authorship page. We acquire the corresponding text description with Whisper~\cite{radford2022robust} from the audio file, and for E-books that do not have corresponding audio files, but with available storyline text, we use OCR algorithms, to directly recognize the text on each page.

\begin{table}[t]
        \centering
        \footnotesize
 \setlength
 \tabcolsep{3pt}
         \begin{tabular}{c|c|ccc}
         \toprule
 Dataset & Style & \#Frames & Avg.Length & \#Categories \\
          \midrule
  PororoSV~\cite{li2019storygan} & Animation & 73,665 & 5 & 9 \\
  FlintstonesSV~\cite{gupta2018imagine} & Animation  & 122,560 & 5 & 7 \\
  DiDeMoSV~\cite{maharana2022storydall} & Real  & 52,905 & 3 & - \\
  VIST~\cite{huang2016visual} & Real  & 145,950 & 5 & - \\ 
  \midrule
  \textbf{StorySalon} & Animation & \textbf{159,778} & \textbf{14} & \textbf{446} \\
          \bottomrule
        \end{tabular}
        \vspace{-2pt}
        \caption{\textbf{Dataset Statistics.} Our StorySalon dataset far exceeds previous story generation datasets in terms of the total number of images, average length, and categories of characters included.}
        \label{tab:Dataset}
        \vspace{-0.4cm}
\end{table}

\begin{table*}[t]
\hspace{5mm}
\begin{minipage}{.34\linewidth}
        \centering
        \footnotesize
 \setlength
 \tabcolsep{3pt}
         \begin{tabular}{c|ccc}
         \toprule
 Model & FID~$\downarrow$ & CLIP-I~$\uparrow$ & CLIP-T~$\uparrow$ \\
          \hline
  GT & - & 1.0 & 0.2668\\ \hline
  SDM & 73.50 & 0.6155 & 0.3218\\
  Prompt-SDM & 67.35 & 0.6272 & \textbf{0.3225}\\
  Finetuned-SDM & 42.01 & 0.6970 & 0.3005\\
  \midrule
  StoryDALL·E & 38.34 & 0.6823 & 0.2366\\
  AR-LDM & 39.55 & 0.6864 & 0.2614\\
  \textbf{StoryGen} & \textbf{33.90} & \textbf{0.7467} & 0.2875\\
          \bottomrule
        \end{tabular}
        \vspace{3pt}
        \caption{
        \textbf{Comparison of automatic metrics} on StorySalon test set.
        Prompt-SDM denotes Stable Diffusion model with cartoon-style-directed prompts and 
        Finetuned-SDM represents a Stable Diffusion model with all parameters fine-tuned on our StorySalon dataset. 
        }
        \label{tab:quantitative}
        \vspace{12pt}
\end{minipage}
\hspace{5mm}
\begin{minipage}{.58\linewidth}
\centering
\footnotesize
\begin{tabular}{ccccccc}
\toprule
\multicolumn{7}{c}{Story Generation} \\ 
\toprule
\multicolumn{1}{c|}{Model} & Align.~$\uparrow$ & Style~$\uparrow$ & Cont.~$\uparrow$ & Char.~$\uparrow$ & \multicolumn{1}{c|}{Qual.~$\uparrow$} & Pref.~$\uparrow$ \\
\hline
\multicolumn{1}{c|}{GT} & 4.04 & 4.66 & 4.41 & 4.54 & \multicolumn{1}{c|}{4.29} & -- \\
\hline
\multicolumn{1}{c|}{SDM} & 3.61 & 2.88 & 2.90 & 2.51 & \multicolumn{1}{c|}{3.74} & 14.05\% \\
\multicolumn{1}{c|}{Prompt-SDM} & 3.39 & 2.56 & 2.68 & 2.10 & \multicolumn{1}{c|}{3.44} & 8.57\% \\
\multicolumn{1}{c|}{StoryGen-S} & 3.50 & 2.73 & 2.81 & 2.21 & \multicolumn{1}{c|}{3.19} & 10.24\% \\
\multicolumn{1}{c|}{\textbf{StoryGen}} & \textbf{3.78} & \textbf{4.79} & \textbf{4.26} & \textbf{4.64} & \multicolumn{1}{c|}{\textbf{3.76}} & \textbf{67.14\%} \\ 
\midrule
\multicolumn{7}{c}{Story Continuation} \\ 
\midrule
\multicolumn{1}{c|}{StoryDALL·E} & 1.18 & 1.55 & 1.20 & 1.14 & \multicolumn{1}{c|}{1.19} & 0.63\% \\
\multicolumn{1}{c|}{AR-LDM} & 2.47 & 2.82 & 2.40 & 1.87 & \multicolumn{1}{c|}{2.54} & 2.50\% \\
\multicolumn{1}{c|}{\textbf{StoryGen}} & \textbf{4.23} & \textbf{4.70} & \textbf{4.35} & \textbf{4.38} & \multicolumn{1}{c|}{\textbf{4.18}} & \textbf{96.87\%} \\ 
\bottomrule
\end{tabular}
\caption{
    \textbf{Comparison results of human evaluation.}
    GT stands for ground truth from the test set. 
    StoryGen-S represents StoryGen without context conditions.
    The abbreviated metrics are Text-image alignment, Style consistency, Content consistency, Character consistency, image quality, and Preference, respectively.
}
\label{tab:human_evaluation}
\end{minipage}
\vspace{-8pt}
\end{table*}

\vspace{2pt}\noindent{\bf Visual-Language Alignment.}
As shown in Figure~\ref{fig:dataset}, for each of the image, 
we can collect two types of text descriptions, {\em e.g.}, story-level narration, and descriptive captions. This is based on our observation that there actually exists a semantic gap between narrative storyline and descriptive text, for example, the same image can be well described as {\em ``The cat is isolated by others, sitting alone in front of a village."} in the story, or {\em ``A black cat sits in front of a number of houses."} as descriptive caption, 
therefore, directly fine-tuning stable diffusion models with story narration may be detrimental to its pre-trained text-image alignment. 
In practice, to get story-level paired image-text samples, 
we align the subtitles with visual frames by using Dynamic Time Warping (DTW) algorithm~\cite{muller2007dynamic}.
To get visual descriptions, we use TextBind~\cite{TextBind} to generate captions for each image, with both the image and the corresponding narrative text as inputs.
At training time, this allows us to substitute the original story with more accurate and descriptive captions.

\vspace{2pt} \noindent{\bf Visual Frame Post-processing.}
In practice, we observe that book pages and borders in images can potentially interfere with our generative model by having story texts printed on them. 
To tackle this, we use an OCR detector to identify text regions in images and an image inpainting model~\cite{SDM} to fill in the text and headshot regions, resulting in more precise image-text pairs that are suitable for model training.

\vspace{2pt} \noindent \textbf{Discussion.}
After the three-step pipeline above, we obtain our StorySalon dataset.
As shown in Table~\ref{tab:Dataset}, our dataset has nearly 160K animation-style images in total with an average length of 14 frames per story, which is conducive to building long-range semantic correspondence.
Finally, we query MiniGPT-4~\cite{zhu2023minigpt} about the main character category of each image in our dataset, like {\em Dog} and {\em Cat}, then count the categories and filter out those appear less than 3 times. Compared with previous datasets with less than 10 characters, our dataset comprises hundreds of character categories, and even more character instances, which provides a data basis for training open-ended visual storytelling models, showing a significantly broader range of visual styles and character appearances over existing datasets.

\section{Experiments}
\label{sec:exp}
In this section, we start by describing our experimental settings,
then compare with other models from three different perspectives: 
image-text alignment, consistency and image quality with subjective human evaluation and quantitative metrics. 
Additionally, we present results for ablation experiments to prove the effectiveness of our proposed modules.

\subsection{Experimental Settings}
\noindent \textbf{Training Details.}
Our model is built on the stable diffusion v1.5 model, 
and trained with a learning rate of $1 \times 10^{-5}$ and a batch size of $256$. 
We begin with a single-frame self-attention pre-training stage, 
which involves 3,000 iterations on 8 NVIDIA RTX3090. 
Next, we incorporate our proposed vision-language context module, 
and train it for 5,000 iterations using a single preceding image-caption pair as context condition, then continue to train it for another 5,000 iterations with multiple image-caption pairs for multi-frame conditioning.
To maintain our model's unconditional denoising ability for classifier-free guidance, we randomly drop the current text and the context image-caption pairs with a probability of 5\% and 15\%, respectively.
During inference, we utilize DDIM~\cite{song2020denoising} with 40 steps of sampling and select the guidance weight $w_v = 7.0$ and $w_t = 3.5$.

\begin{figure*}[htb]
  \centering
  \includegraphics[width=\textwidth]{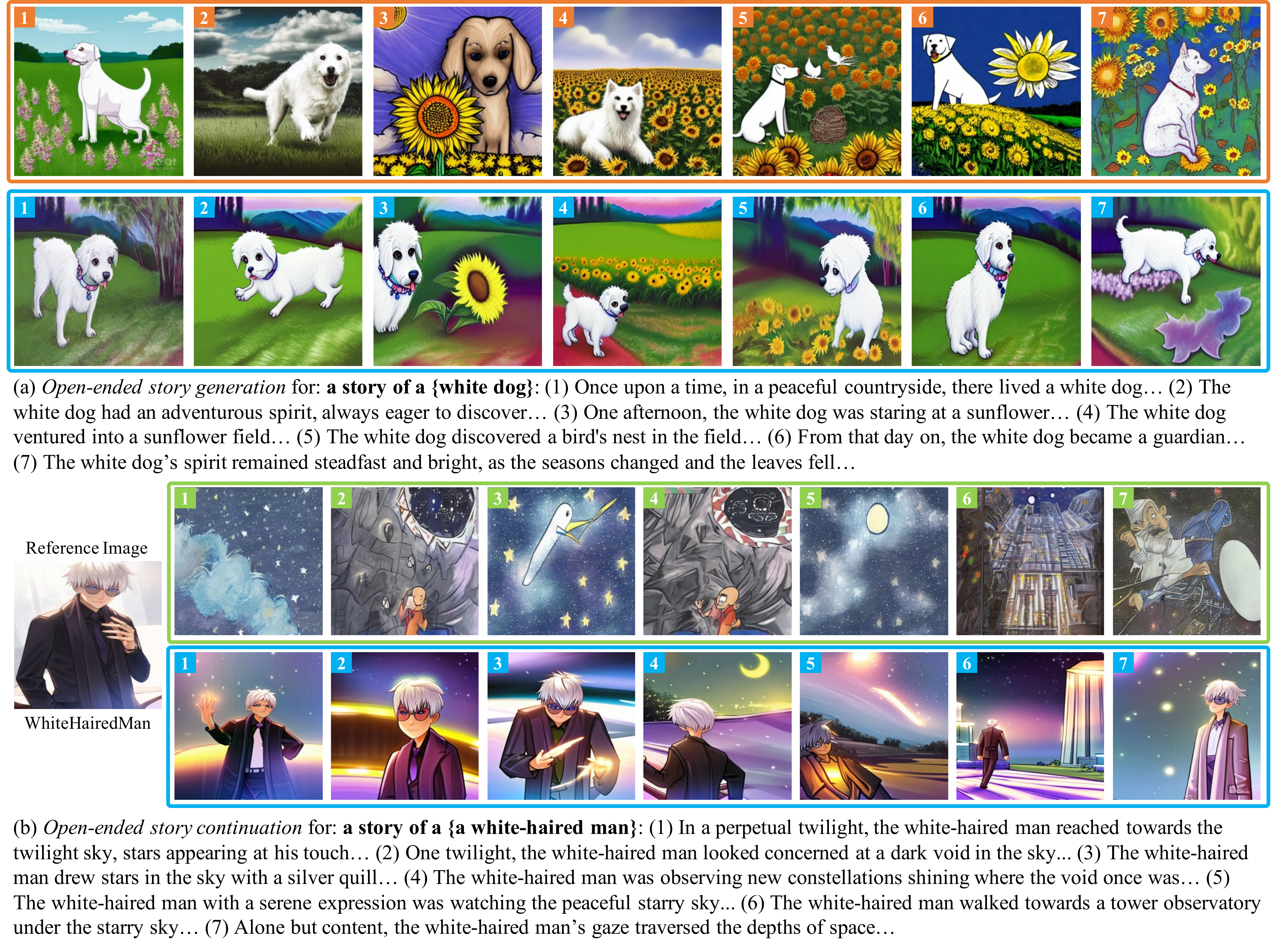} \\
  \vspace{-0.25cm}
  \caption{
  \textbf{Qualitative Comparison with other methods}. 
  The image sequences in \textcolor{orange}{orange}, \textcolor{color_green}{green}, and \textcolor{cyan}{blue} boxes are generated by \textcolor{orange}{Prompt-SDM}, \textcolor{color_green}{AR-LDM} and \textcolor{cyan}{StoryGen} respectively. 
  Our synthesis results exhibit impressive performance superiority in terms of style, content and character consistency, text-image alignment, and image quality. 
  Please refer to the Appendix for more qualitative results.
  } 
  \vspace{-0.45cm}
 \label{fig:qualitative}
\end{figure*}

\vspace{2pt} \noindent \textbf{Baselines.}
We consider two scenarios of our proposed open-ended storytelling task, namely, story generation and story continuation. 
For \textbf{story generation}, we need the model to be able to generate a complete visual story only based on a given storyline.
So we present a comparison with Stable Diffusion Model (\textbf{SDM}) and \textbf{Prompt-SDM}, which conditions on an additional cartoon-style-directed prompt {\em ``A cartoon style image"}. 
For \textbf{story continuation}, the first frame or the main character is given, and the model is expected to generate coherent images based on the storyline.
In this scenario, we compare our model with two closed-set story continuation models: namely, \textbf{StoryDALL·E}~\cite{maharana2022storydall} and \textbf{AR-LDM}~\cite{pan2022synthesizing} re-trained on our StorySalon dataset.

\vspace{2pt} \noindent \textbf{Automatic Metrics.}
To evaluate the quality of generated image sequences, 
we adopt three widely-used metrics, including Fréchet Inception Distance score (FID)~\cite{heusel2017gans}, CLIP image-image similarity (CLIP-I), and CLIP text-image similarity (CLIP-T).
Notably, in order to avoid the impact of randomness in synthesis quality, we utilize a CLIP-based scoring function trained exclusively on text-to-image generated images, namely, 
PickScore~\cite{PickScore}, to automatically select the generated images with better quality.
Each chosen image is selected from a pool of 10 candidates.

\subsection{Quantitative Evaluation Results}
We compare our StoryGen model with other baselines on StorySalon test set, which contains 5\% of total data 
(nearly 7K pairs).
Each contains a current prompt and the image-text context of the previous frame. 
The models are expected to generate the current frame based on given conditions.

The quantitative results in Table~\ref{tab:quantitative} demonstrate that our StoryGen model exhibits significant performance improvement in terms of FID score and CLIP-I similarity compared to existing models, while maintaining comparable CLIP-T similarity.
This confirms that our model can effectively exploit contextual information, thus generating animation-style visual stories based on the given storyline.
Notably, CLIP trained on natural images tends to have an understanding bias towards animation-style images, and the slight decline in CLIP-T is an inevitable result of the conflict between text condition and newly introduced image condition.

\subsection{Human Evaluation Results}
Considering that the above metrics may not reflect the quality of the generated stories accurately, and there is no standardized metric for evaluating the consistency within the visual story, we further include human evaluation for comparison of image-text alignment, image style, story consistency, character consistency and synthesis quality.

For the two scenarios mentioned above, we respectively conduct two types of human evaluation to assess the quality of generated visual stories. 
To mitigate bias, participants are unaware of the type of storybooks they are evaluating.
Concretely, we prompt GPT-4 to produce multiple storylines for both test modes, and for story continuation, we search the Internet for multiple characters that have never appeared in our dataset. Then we utilize our StoryGen along with other baselines to generate corresponding sequences of images.
\begin{figure*}[htb]
  \centering
  \includegraphics[width=\textwidth]{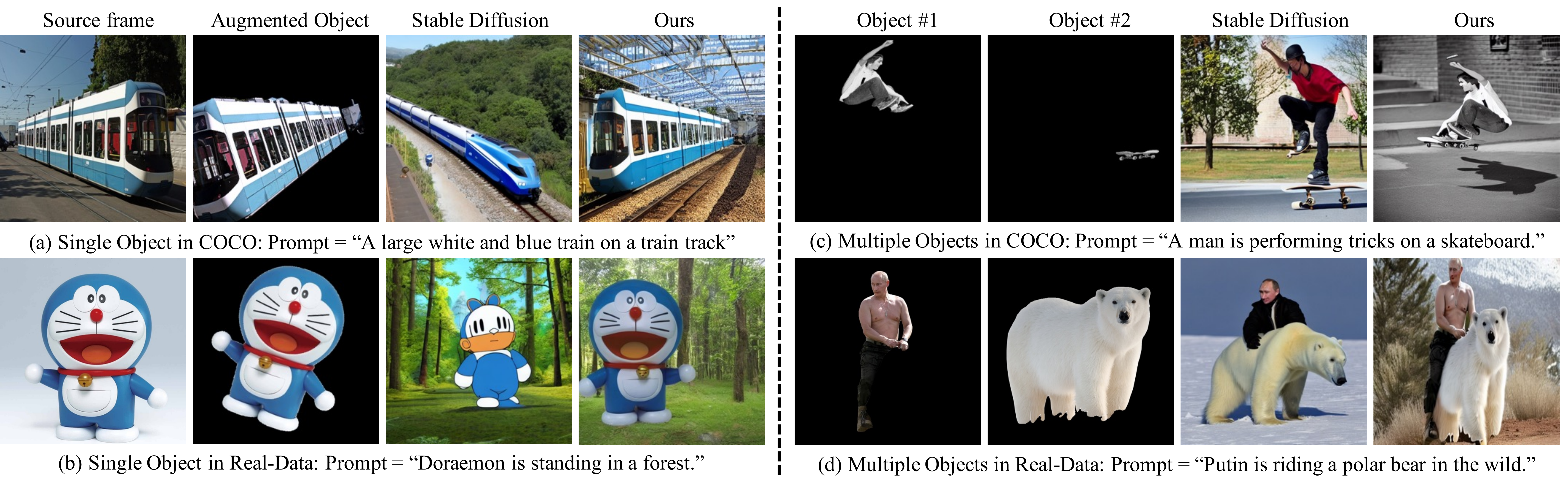} \\
  \vspace{-6pt}
  \caption{
  \textbf{Ablation studies on consistency}.
  We incorporate our proposed Visual-Language Context Module into a pre-trained SDM, and train it on MS-COCO~\cite{lin2014microsoft} with other parameters frozen.
  The content consistency of single-object and multi-object generation on COCO and real data has demonstrated the effectiveness of our module.
  Please refer to the Appendix for experiment details and quantitative results.
  }
 \label{fig:ablation}
 \vspace{-6pt}
\end{figure*} 

\vspace{2pt} \noindent \textbf{Protocol-I}. 
We randomly select
an equal number of samples from the generated results of our StoryGen and other baselines.
Each time we randomly sample a visual story from these sources, and participants are then invited to rate the sample with a score ranging from 1 to 5, taking into account text-image alignment, style consistency, content consistency, character consistency and image quality.
Higher scores indicate better samples.
We also evaluate the same number of samples from StorySalon test set as a reference.

\vspace{2pt} \noindent \textbf{Protocol-II}. 
Each time we randomly sample a storyline and its corresponding visual storybooks generated by StoryGen and other methods. 
Participants are invited to select their preferred generated result among these different image sequences of the same storyline.

\vspace{2pt}\noindent \textbf{Results}. 
The results of human evaluation presented in Table~\ref{tab:human_evaluation} illustrate that our StoryGen model demonstrates excellent performance in overall score, especially in terms of consistency and quality. 
This indicates that our model can generate coherent image sequences that are highly consistent with given text prompts and visual-language contexts.

\subsection{Qualitative Results}
\label{sec:qualitative}
In Figure~\ref{fig:qualitative}, 
we present visualization results of both open-ended visual story generation and visual story continuation, showing that our StoryGen can generate visual stories with a broad vocabulary, while maintaining content coherence and character consistency throughout the narrative, whereas other methods fail to do so.
Moreover, our model can stably maintain the animation style of generated images, which satisfies the requirements of visual storytelling for children. More results can be found in the supplementary material.

\subsection{Ablation Studies}
In order to demonstrate the effectiveness of our proposed modules, 
we conduct ablation studies from both quantitative metrics and qualitative visualization.

\vspace{2pt} \noindent {\bf On Variants of StoryGen.}
We evaluate the performance of multiple model variants on the StorySalon test set, including (i) our model without the context module, marked as \textbf{StoryGen-Single}, which solely fine-tunes the self-attention layers on our dataset.
(ii) our model with context features encoded by the VAE of SDM as context condition, without text-guided diffusion process, denoted as \textbf{StoryGen-VAE};
(iii) our model with CLIP image embedding as context condition (\textbf{StoryGen-CLIP}); 
(iv) our model with context features extracted by BLIP image encoder (\textbf{StoryGen-BLIP});
(v) our model with naive denoising features at \textbf{L}arge-scale diffusion \textbf{T}imestep, satisfying $t' = t$, as condition (\textbf{StoryGen-LT});
and (vi) our full model (\textbf{StoryGen}).
We also employ PickScore to filter generation results of all these models.
The findings presented in Table~\ref{tab:ablation} illustrate the inclusion of our context module can significantly improve the model performance, in terms of CLIP-I and FID.
As for the slight inferiority in CLIP-T, we have claimed above that this is due to the understanding bias towards animation-style images for CLIP trained on natural images.





\vspace{2pt} \noindent{\bf Qualitative Visualization.}
As mentioned above, consistency is a crucial factor in visual story generation. 
We hope to more intuitively demonstrate that our proposed context module can accurately capture the image content of the previous frame.
To this end, we incorporate our context module into SDM and train it from scratch on the MS-COCO~\cite{lin2014microsoft} with other parameters frozen.
Specifically, we crop the object and perform data augmentations such as translation and rotation to use it as image condition. 
The category of the cropped object is used as its corresponding text, and the caption of the original image serves as the text prompt. 
We expect the model to reconstruct the original image relying on the conditions above, which enables the context module to learn how to leverage the previous image.
As shown in Figure~\ref{fig:ablation}, our model can make full use of the objects in the reference frame and generate new images that are consistent with them, while SDM fails to do so. 
In addition, this can also be transferred to any real-world reference image, which strongly illustrates the robustness and capability of our context module to assist diffusion models in generating images based on any given object.

\begin{table}[t]
        \centering
        \small
 \setlength
 \tabcolsep{3pt}
         \begin{tabular}{c|ccc}
         \toprule
 Model & FID~$\downarrow$ & CLIP-I~$\uparrow$ & CLIP-T~$\uparrow$\\
          \hline
  StoryGen-Single & 38.81  & 0.6869 & \textbf{0.3140} \\
  StoryGen-VAE & 36.98 & 0.6846 & 0.3061 \\
  StoryGen-CLIP & 36.66 & 0.6934 & \textbf{0.3140} \\
  StoryGen-BLIP & 34.78 & 0.7026 & 0.2838 \\
  StoryGen-LT & 36.41 & 0.7141 & 0.3025 \\
  \textbf{StoryGen} & \textbf{33.90} & \textbf{0.7467} & 0.2875 \\
          \bottomrule
        \end{tabular}
        \vspace{-3pt}
        \caption{\textbf{Ablation studies} on Visual-Language Context Module. 
        }
        \vspace{-9pt}
        \label{tab:ablation}
\end{table}


\section{Conclusion}
\vspace{-0.1cm}
In this paper, 
we consider an interesting, yet challenging task, 
termed as {\em open-ended visual storytelling}, 
which involves generating a sequence of images that tell a coherent visual story based on the given storyline. 
Our proposed learning-based \textbf{StoryGen} model can take input from the preceding image-caption context along with the text prompt to generate coherent image sequences in an auto-regressive manner,
{\em i.e.}, without test-time optimization.
On the data side, we establish a data processing pipeline to collect a large-scale dataset named \textbf{StorySalon} that comprises storybooks with diverse characters, storylines, and artistic styles sourced from videos and E-books. 
Extensive human evaluation and quantitative comparison have illustrated that our proposed model substantially outperforms existing models, from the perspective of image quality, content coherence, character consistency, and visual-language alignment.

\section*{Acknowledgments}
This work is supported by National Key R\&D Program of China (No. 2022ZD0161400), 
National Natural Science Foundation of China (62271308), STCSM (22511105700, 22DZ2229005), 111 plan (BP0719010), and State Key Laboratory of UHD Video and Audio Production and Presentation.

{
    \small
    \bibliographystyle{ieeenat_fullname}
    \bibliography{main}
}


\onecolumn
{
    \centering
    \Large
    \textbf{Intelligent Grimm - Open-ended Visual Storytelling via Latent Diffusion Models}\\
    \vspace{0.5em}Supplementary Material \\
    \vspace{1.0em}
}
\setcounter{page}{1}
\appendix
{
  \hypersetup{linkcolor=black}
  \tableofcontents
}
\clearpage

\section{Preliminaries on Diffusion Models}
\label{sec:pre}
Diffusion models are a type of generative models that undergo a denoising process, converting input noise into meaningful data samples.
Diffusion models comprise a forward diffusion process that incorporates Gaussian noise into an image sample $\mathbf{x}_0$, accomplished via a Markov process over $T$ steps. If we denote the noisy image at step $t$ as $\mathbf{x}_t$, the transition function $q(\mathbf{x}_t|\mathbf{x}_{t-1})$ connecting $\mathbf{x}_{t-1}$ and $\mathbf{x}_t$ can be expressed as follows: 
\vspace{-9pt}
\begin{align*}
    q(\mathbf{x}_t \vert \mathbf{x}_{t-1}) = \mathcal{N}(\mathbf{x}_t; \sqrt{1 - \beta_t} \mathbf{x}_{t-1}, \beta_t\mathbf{I}) \quad
q(\mathbf{x}_{1:T} \vert \mathbf{x}_0) = \prod^T_{t=1} q(\mathbf{x}_t \vert \mathbf{x}_{t-1})
\end{align*}
where $\beta_t \in (0, 1)$ is the variance schedule controlling the step size.

Using Gaussian distribution property and reparameterization, if we define $\alpha_t = 1 - \beta_t$ and $\bar{\alpha}_t = \prod_{i=1}^t \alpha_i$, we can write the equation above as follows:
\begin{align*}
\label{eqn:forwardG}
q(\mathbf{x}_t \vert \mathbf{x}_0) = \mathcal{N}(\mathbf{x}_t; \sqrt{\bar{\alpha}_t} \mathbf{x}_0, (1 - \bar{\alpha}_t)\mathbf{I})
\end{align*}

Diffusion models also comprise a reverse diffusion process that learns to restore the initial image sample from noise.
A UNet-based model~\cite{ronneberger2015u} is utilized in the diffusion model to learn the reverse diffusion process $p_\theta$.
The process $p_\theta$ can be expressed using the following equation.
\begin{align*}
p_\theta(\mathbf{x}_{0:T}) = p(\mathbf{x}_T) \prod^T_{t=1} p_\theta(\mathbf{x}_{t-1} \vert \mathbf{x}_t) \quad
p_\theta(\mathbf{x}_{t-1} \vert \mathbf{x}_t) = \mathcal{N}(\mathbf{x}_{t-1}; \boldsymbol{\mu}_\theta(\mathbf{x}_t, t), \boldsymbol{\Sigma}_\theta(\mathbf{x}_t, t))
\end{align*}
where $\boldsymbol{\mu}_\theta$ is the predicted Gaussian distribution mean value. 

As we compute the loss function by taking the mean absolute error of the noise term $\boldsymbol{\epsilon}_\theta$ into account, we can express the mean value $\boldsymbol{\mu}_\theta$ in terms of the noise term $\boldsymbol{\epsilon}_\theta$ as follows:
\begin{align*}
\boldsymbol{\mu}_\theta(\mathbf{x}_t, t) = \frac{1}{\sqrt{\alpha_t}} \Big( \mathbf{x}_t - \frac{1 - \alpha_t}{\sqrt{1 - \bar{\alpha}_t}} \boldsymbol{\epsilon}_\theta(\mathbf{x}_t, t) \Big)
\end{align*}

Therefore, the objective can be written as:
\begin{align*}
\mathcal{L}_t = \mathbb{E}_{t \sim [1, T], \mathbf{x}_0, \boldsymbol{\epsilon}_t} \Big[\|\boldsymbol{\epsilon}_t - \boldsymbol{\epsilon}_\theta(\mathbf{x}_t, t)\|^2 \Big]
\end{align*}

\section{Further Architecture Details}
\label{sec:farch}
In this section, we will provide a more comprehensive illustration about more design details of our model.

\subsection{Parameter Sharing Strategy}
Initially, we will discuss the strategy of parameter sharing between the standard diffusion UNet and our StoryGen.
As illustrated in Figure 2 of the main paper, the standard diffusion UNet is exploited in {\em Preceding Feature Extraction} to extract diffusion context features, and StoryGen is utilized in {\em Current Frame Generation} to generate new frames with consistency and coherence.


Specifically,
the parameters of the corresponding attention layers are shared between the standard diffusion UNet and our StoryGen,
including self-attention layers and text cross-attention layers.
In practise, the standard diffusion UNet here is a modified version of our StoryGen, without the image cross-attention layers, and all other parameters are shared.
This design allows the UNet to extract contextual diffusion features within the same latent space as StoryGen.




\subsection{Multi-frame Condition Strategy}
As illustrated in Figure 2 and Section 3.2 of the main paper, when dealing with multiple preceding image-caption pairs, we add more noise (corresponding to a larger diffusion timestep $t'$) to reference frames with longer temporal distances to the current frame.
Such design effectively serves two purposes,
{\em first}, it is based on the observation that frames with longer temporal distances will incur less effect on the generation of the current frame;
{\em second}, the different noise level also serves as positional encoding, allowing for the differentiation of temporal order, which enables us to directly concatenate these diffusion context features.

Specifically, we use $t$ to represent the diffusion timestep of the current image $\hat{\mathcal{I}}_{k}$, and use $t'_{j}$ to represent the diffusion timestep of the preceding image-text pair $(\hat{\mathcal{I}}_{j},\mathcal{T}_{j})$.
When generating image $\hat{\mathcal{I}}_{k}$, 
we use $t'_{k-1} =t / 10$ for $(\hat{\mathcal{I}}_{k-1},\mathcal{T}_{k-1})$ pair, $t'_{k-2} =2t/10$ for $(\hat{\mathcal{I}}_{k-2},\mathcal{T}_{k-2})$ pair, and so on.
In summary, the diffusion timestep $t'_{k-i}$ for $(\hat{\mathcal{I}}_{k-i},\mathcal{T}_{k-i})$ pair will follow a $t'_{k-i} =i * t/10$ rule.


\subsection{Two-stage Training Strategy}
As illustrated in Section 3.3 of the main paper, we exploit a two-stage training strategy, including single-frame pre-training and multiple-frame fine-tuning. 
In \textbf{single-frame pre-training stage}, we do not introduce additional image cross-attention layers, and only train the self-attention layers in standard SDM~\cite{SDM} on our dataset. The goal of this stage is to train the model for single-frame generation in the style of storybooks.
In \textbf{multiple-frame fine-tuning stage}, we train the additional image cross-attention layers in vision-language context module on our dataset, with all other parameters frozen.
We first train image cross-attention layers with a single preceding image-caption pair, and then continue with multiple image-caption pairs.
Consequently, StoryGen acquires the capability to condition on multiple preceding image-caption pairs and generate image sequences in an auto-regressive manner.
Throughout the entire two-stage training, the text cross-attention layers remain frozen, preserving the vision-language alignment inherited from the pre-trained stable diffusion models.

\begin{figure}[t]
  \centering
  \includegraphics[width=0.95\textwidth]{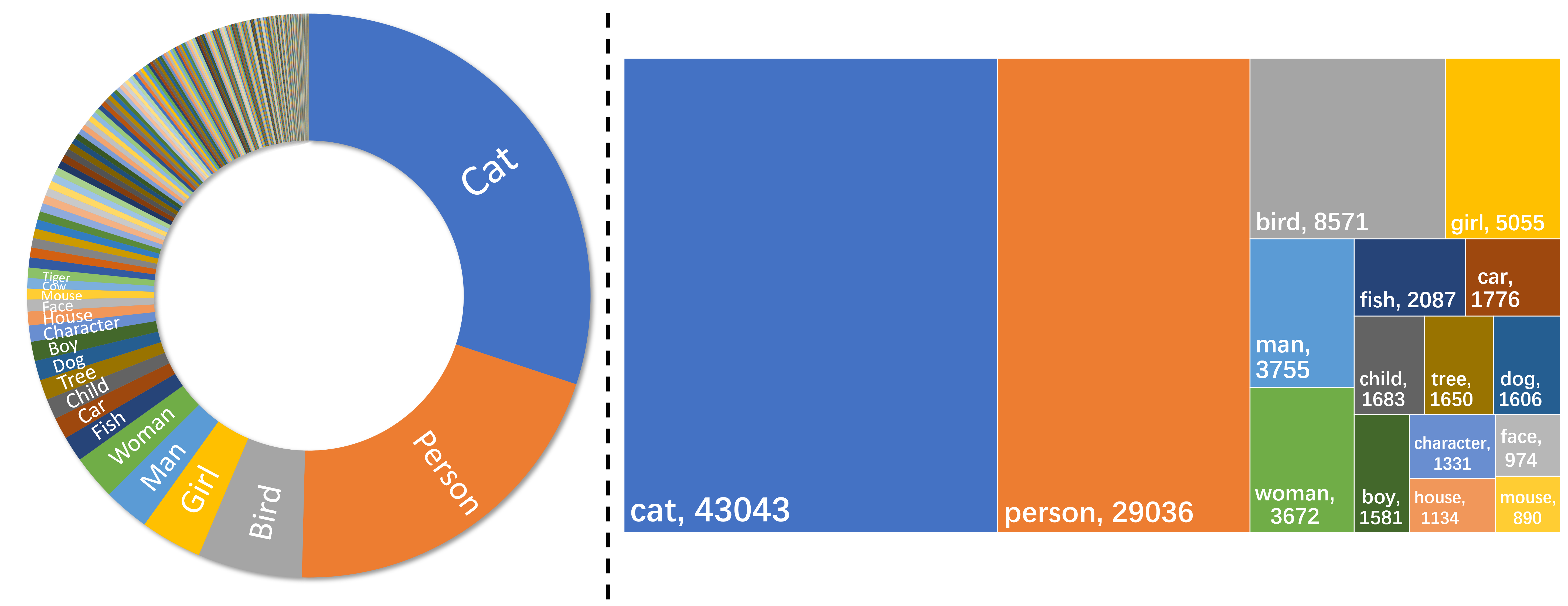} 
  \caption{ \textbf{Dataset Statistics Results.} 
  \textbf{Left:} Distribution of text-image pairs classified by the main character categories in our collected StorySalon dataset.
  \textbf{Right:} The top 16 character categories and corresponding numbers in StorySalon, cover a wide range of character types.
  }
 \label{fig:stat}
 \vspace{-6pt}
\end{figure}

\section{Dataset Details}
In Section~\ref{sec:datasource}, we present additional details about the data sources of our StorySalon dataset. Subsequently, we show the detailed statistics of our dataset in Section~\ref{sec:stat}.

\subsection{Data Sources}
\label{sec:datasource}

Our StorySalon dataset mainly comprises of two components, 
{\em e.g.}, online videos collected from YouTube, and open-source E-books collected from six online libraries. For online videos, we download a large number of videos and corresponding subtitles from YouTube, by querying keywords related to story-telling for children, for instance, {\em storytime}. For open-source E-books, we collect E-books~(partially with corresponding audios available) from six open-source online libraries which are all registered under the Creative Commons 4.0 International Attribution (CC BY 4.0) license. These online libraries have consistently dedicated themselves to assisting children in underdeveloped regions.
We extend our appreciation for their ongoing endeavors and contributions.
Specifically, these open-source online libraries include: 
\vspace{0.2cm}
\begin{itemize}
  \setlength\itemsep{0.5em}
    \item \textbf{African Storybook.} \url{https://africanstorybook.org};
    \item \textbf{Bloom Library.} \url{https://bloomlibrary.org/};
    \item \textbf{Book Dash.} \url{https://bookdash.org/};
    \item \textbf{Global Digital Library.} \url{https://digitallibrary.io/topic/library-books/};
    \item \textbf{Room to Read.} \url{https://literacycloud.org/};
    \item \textbf{Digital Library of Illustrated Storybooks.} \url{https://storyweaver.org.in/en}.
\end{itemize}

\subsection{Dataset Statistics}
\label{sec:stat}


Our StorySalon dataset comprises a total of $11,280$ storybooks and $159,778$ text-image pairs, 
with approximately 160K animation-style images, 
averaging 14 frames per narrative, 
as demonstrated in Table 1 of the main paper.
We divide the dataset into train and test sets following a $9:1$ ratio.
Both the video and E-book components are randomly split into train and test sets according to this proportion.




To categorize the characters in storybooks, we use 
MiniGPT-4~\cite{zhu2023minigpt} to infer the predominant character category in each image of our dataset, 
such as {\em Dog} and {\em Cat}.
Subsequently, we count these categories and exclude those occurring fewer than three times.
The distribution of these character categories is depicted in Figure~\ref{fig:stat}.
In contrast to preceding datasets featuring fewer than ten characters, our dataset encompasses hundreds of character categories, with rich appearances.
Consequently, StorySalon offers the data foundation for training open-ended visual storytelling models.

\section{Consistency Ablation on COCO}
In Section~\ref{sec:COCOdetail}, we will give a brief illustration on further details about our qualitative consistency ablation experiment on COCO~\cite{lin2014microsoft}.
Subsequently, in Section~\ref{sec:COCOquan}, we will design a new quantitative ablation experiment on COCO, and present additional results.
Finally, in Section~\ref{sec:COCOqua}, we will provide more visualization results of this consistency ablation on COCO.

\subsection{Experiment Details}
\label{sec:COCOdetail}

\vspace{3pt} \noindent {\bf Motivation.}
This experiment serves as an ablation study, designed to show StoryGen's ability in utilizing image conditions, and preserving visual details. Specifically, at training time, we train StoryGen on images from COCO datasets, through a self-supervised learning, by reconstructing the input image, with the image caption as text prompt, and cropped objects as reference. At inference time, the model enables to directly generate images with cropped objects as reference.



\begin{figure*}[t]
  \centering
  \includegraphics[width=\textwidth]{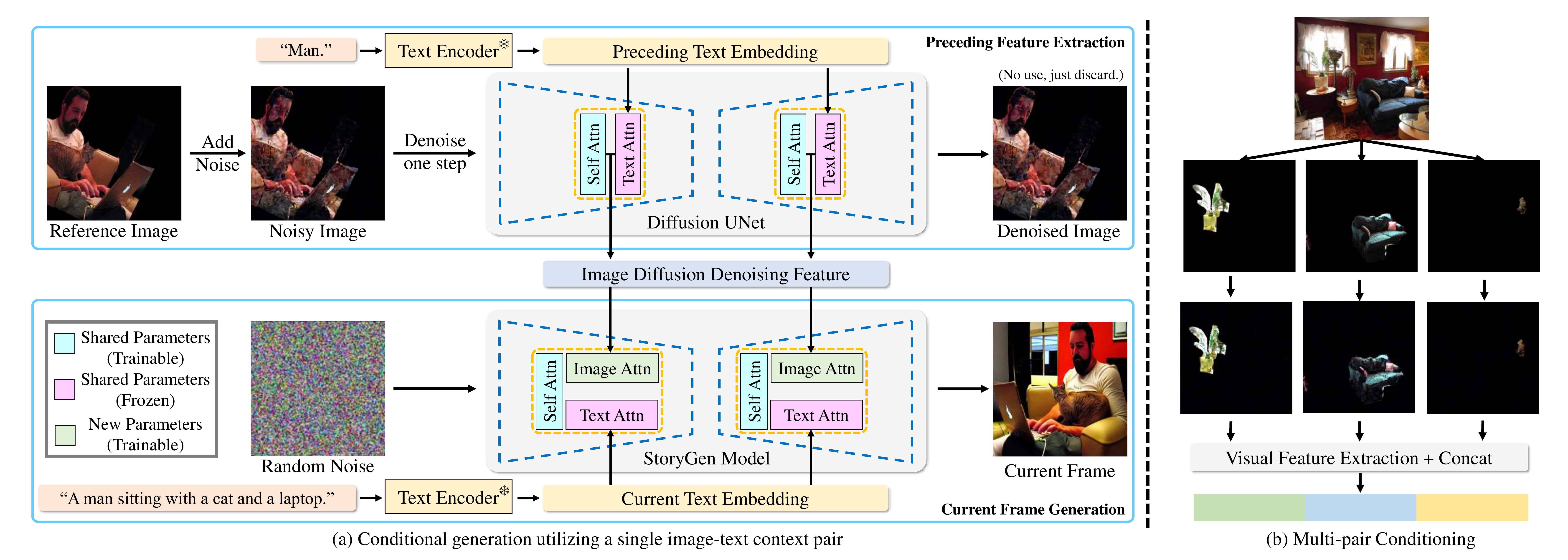} \\
  \vspace{-0.3cm}
  \caption{
  \textbf{Architecture Overview}. 
  (a) Our StoryGen model utilizes current text prompt and previous visual-language contexts as conditions to generate an image.
  (b) In case of multiple image-text context pairs, the multiple features can be directly concatenated to serve as context conditions.
  }
 \label{fig:COCOarch}
 \vspace{-0.4cm}
\end{figure*}

\vspace{3pt} \noindent {\bf Experiment Settings.}
Our experiments on COCO include two scenarios: conditional generation utilizing a single image-text context pair, and alternatively, employing multiple image-text context pairs,
as shown in Figure~\ref{fig:COCOarch}.

\vspace{5pt}\noindent \textbf{Training on  single image-text context pair.}
In this case, we randomly select an image-text pair from the COCO dataset. Initially, we extract all objects with their respective masks, collage them together, and then apply data augmentation, such as translation and rotation to create a composite reference image.  The categories of the extracted objects serve as the reference text prompt candidates, while the captions of the initial image are employed as the text prompt candidates. If multiple candidates for the text prompt or reference text prompt exist, we will randomly choose one.
We expect the model to reconstruct the original image based on these conditions, which enables the context module to learn how to leverage the given reference objects. 
We fine-tune our StoryGen on the train set of COCO2017 for $5,000$ iterations. We only fine-tune the additional image cross-attention layers, and keep self-attention layers and text cross-attention layers frozen.

\vspace{5pt}\noindent \textbf{Training on multiple image-text context pairs.} 
In this case, instead of collaging the objects from an image together,
we individually apply data augmentation to these objects, thereby generating several reference images for a single image-text pair. 
The categories of these objects are consistently selected as the corresponding reference text prompts.
Note that, we employ same diffusion noise scales across these multiple reference images, given that they lack a temporal sequence and hold equal significance.
Taking the model pre-trained for single image-text pair, 
we continue to fine-tune the image cross-attention layers for another $5,000$ iterations, with all other parameters frozen.

\subsection{Quantitative Results}
\label{sec:COCOquan}
To measure consistency, we compute the similarity between the generated image and reference image with a pre-trained DINO~\cite{Mathilde21} model.
Specifically, we compare our StoryGen and variants with the original Stable Diffusion model on the validation set of COCO2017.
Notably, the only difference between original SDM and our StoryGen here, is that StoryGen is augmented with additional image cross-attention layers trained on COCO train set, and all other parameters remain identical between the models. 
Thus, we utilize StoryGen to synthesize the original images with the reference objects, reference text prompts, and image captions, in both single and multiple image-text context pair scenarios;
and we also exploit the original SDM to generate images with the image captions of current frames.
For every image in COCO validation set, ten candidate images are generated.
We use PickScore~\cite{PickScore} to automatically identify those with better visual quality, then calculate the average DINO feature similarity between the ground truth images and the generated images for both StoryGen and original Stable Diffusion models. 

The quantitative results in Table~\ref{tab:COCOquan} demonstrate that our StoryGen model exhibits significant performance improvement in terms of consistency between the generated image and given references. Both StoryGen and StoryGen~(Multiple) outperform the standard SDM.
Moreover, compared to utilizing CLIP or BLIP features as visual conditions, our StoryGen model using diffusion-denoising features as condition demonstrates significant performance advantage, showing its effectiveness for retaining visual details from the reference image.


\begin{table}[!htb]
        \centering
        \small
 \setlength
 \tabcolsep{3pt}
         \begin{tabular}{c|ccccc}
         \toprule
 Model & SDM & StoryGen-CLIP & StoryGen-BLIP & StoryGen & StoryGen~(Multiple)\\
          \midrule
  Consistency Score~$\uparrow$ & 0.4804 & 0.5103 & 0.5475 & 0.7076 & \textbf{0.7317} \\
          \bottomrule
        \end{tabular}
        \vspace{-3pt}
        \caption{Quantitative results on measuring consistency. 
        SDM stands for standard stable diffusion models.
        StoryGen-CLIP and StoryGen-BLIP represent StoryGen with features extracted by CLIP and BLIP image encoder as context condition, as stated in our ablation study.
        StoryGen and StoryGen~(Multiple) stand for StoryGen in single image-text context pair scenario and multiple image-text context pairs scenario, respectively. Consistency Score stands for the average DINO feature similarity, and the higher score yields better results.
        }
        \vspace{-9pt}
        \label{tab:COCOquan}
\end{table}

\subsection{Qualitative Results}
\label{sec:COCOqua}
We provide more visualization samples of our consistency ablation on COCO in this section. 
The results of StoryGen are synthesized with the reference objects, reference text prompts, and image captions, in both single and multiple image-text context pair scenarios.
As the original SDM cannot utilize reference images to exploit contextual visual information, the results of SDM are generated with the image captions only.
The visualization results are all selected from the generated samples on the validation set of COCO2017.


\vspace{5pt}\noindent \textbf{Single-pair COCO Visualization.} 
The qualitative results of our StoryGen on COCO, in single image-text context pair scenario, are depicted in Figure~\ref{fig:COCOS}.
Compared with the results of SDM, our model demonstrates obvious consistency in its generation results.
Despite SDM can generate images that satisfy the text prompts, 
{\em i.e.}, good image-text alignment, the generated images are unable to maintain consistency with the reference image.

\vspace{5pt}\noindent \textbf{Multi-pair COCO Visualization.}  
The qualitative results of our StoryGen on COCO, in multiple image-text context pairs scenario, are depicted in Figure~\ref{fig:COCOM}.
Despite multiple context pairs lead to more potential compositions, 
StoryGen is still able to generate high-quality images with the given reference objects,
maintaining strong object consistency and semantic coherence,
showing the ability of our architecture to exploit reference images for generation.

\begin{figure*}[b]
  \centering
  \vspace{-0.3cm}
  \includegraphics[width=\textwidth]{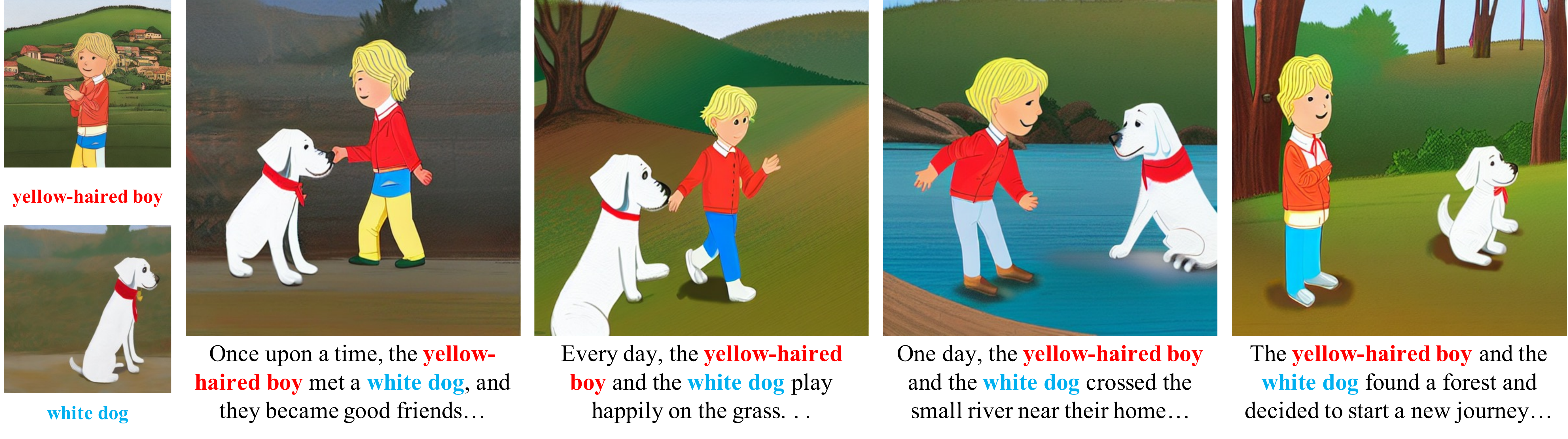} \\
  \vspace{-4pt}
  \caption{
  \textbf{Example of Multi-object Story Continuation}.
  }
 \label{fig:multi_condition_generation}
 \vspace{-8pt}
\end{figure*} 

\begin{figure*}[htb]
  \centering
  \includegraphics[width=\textwidth]{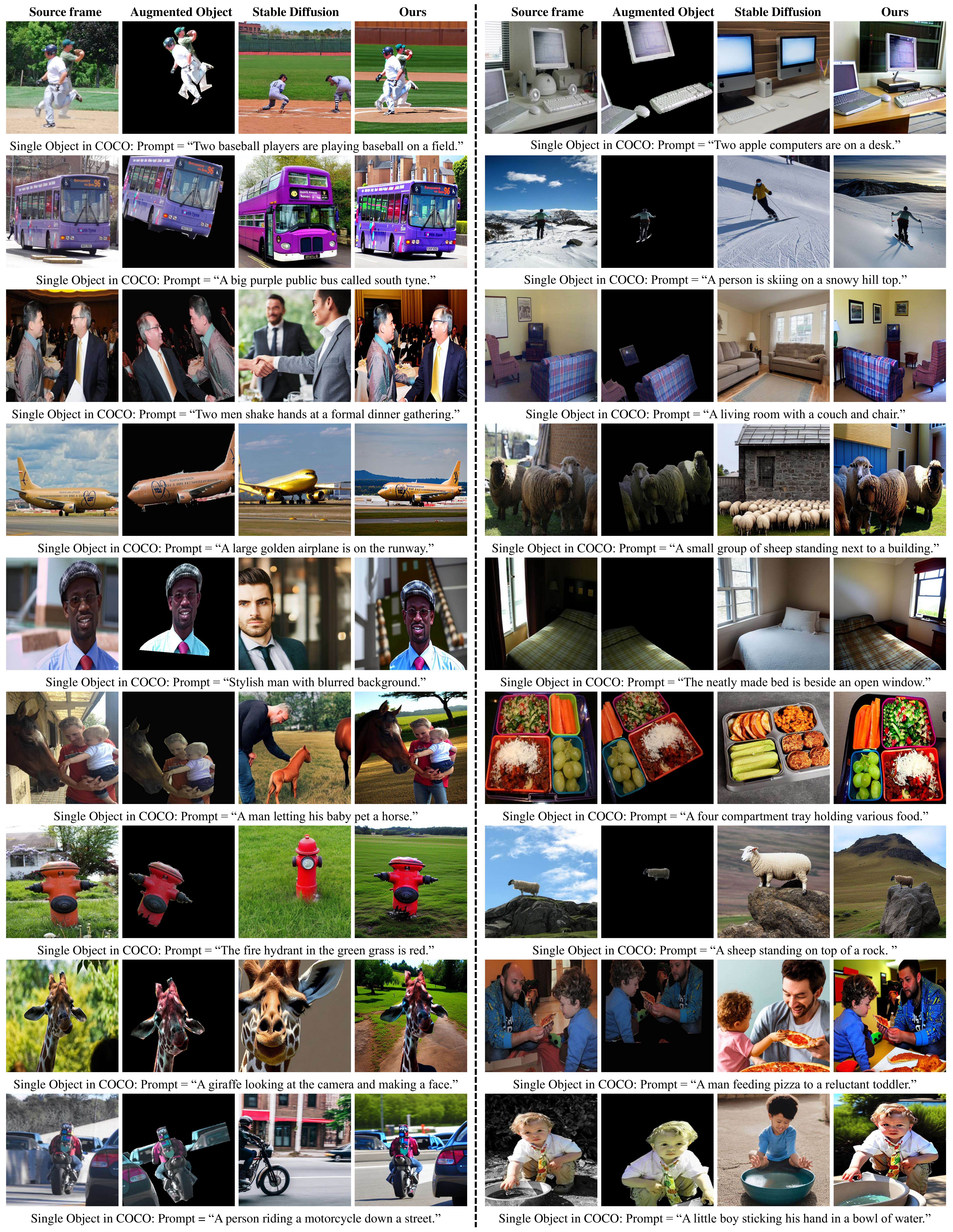} \\
  \vspace{-0.3cm}
  \caption{
  \textbf{Visualization results of StoryGen on COCO in single image-text context pair scenario. }
  }
 \label{fig:COCOS}
\end{figure*}

\begin{figure*}[htb]
  \centering
  \includegraphics[width=\textwidth]{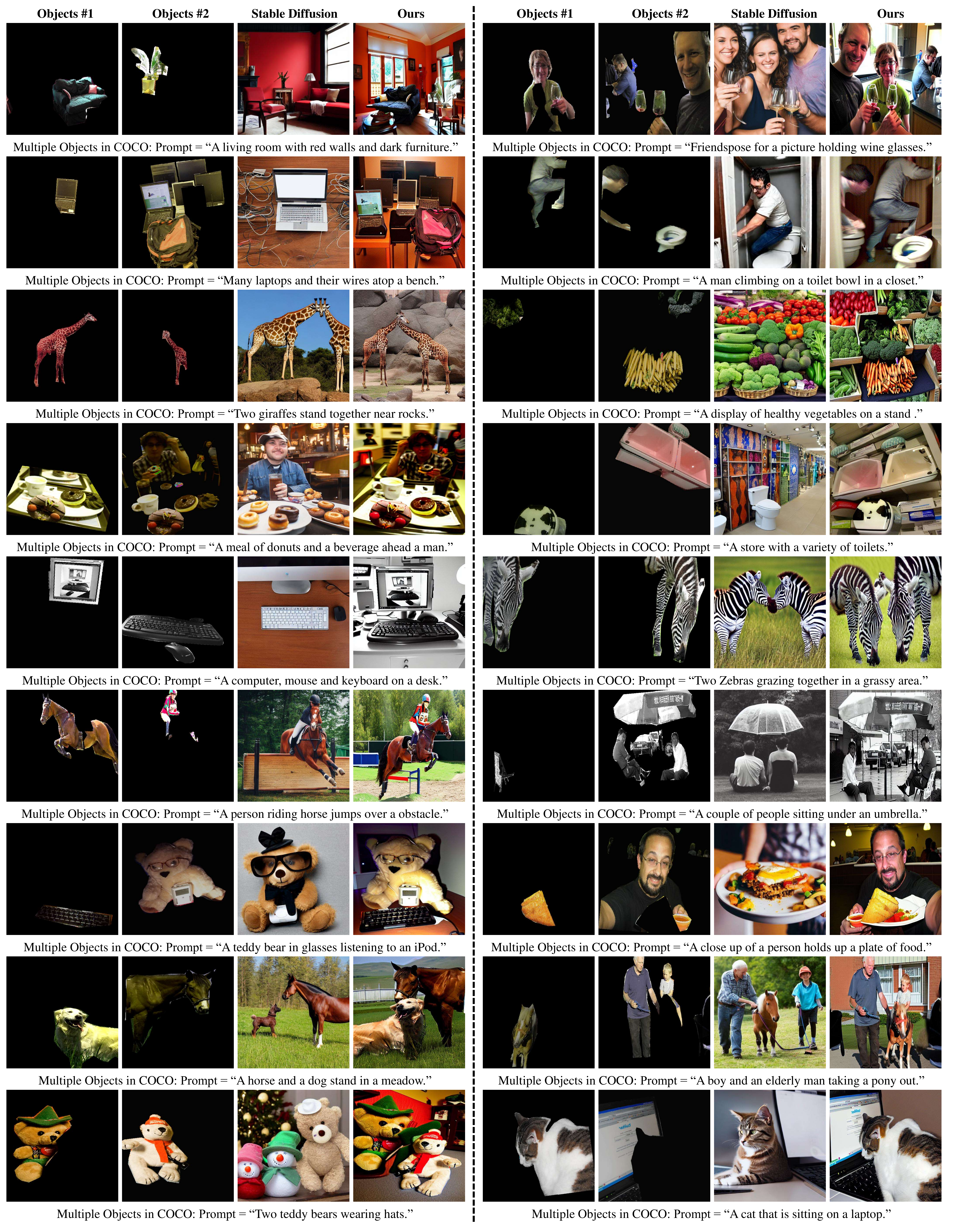} \\
  \vspace{-0.3cm}
  \caption{
  \textbf{Visualization results of StoryGen on COCO in multiple image-text context pairs scenario.}
  }
 \label{fig:COCOM}
\end{figure*}

\section{Broader Impacts}

Our storytelling model also has some positive impacts on the industry of creation and education: The widespread application of our visual storytelling model has the potential to inspire creators and artists to create a large number of visual storybooks rich in basic knowledge, which will have a profound impact on children's early education, as demonstrated by related work in psychology. Our work draws inspiration from those open-source online libraries assisting children in underdeveloped regions,
can potentially assist artists in creating educational storybooks tailored to these young learners.

\section{Limitations}
The principal constraint of our storytelling model lies in the selection of stable diffusion models as its foundational architecture.
Stable diffusion models are known to grapple with significant issues, notably the generation of images with inaccuracies in limb counts (such as legs, arms, or fingers) and decreased quality in the synthesis of images with multiple objects.
Regrettably, our storytelling model inherits these limitations from the stable diffusion model. We anticipate addressing these shortcomings in future research endeavours by considering the adoption of more robust architectures, such as DALL-E 3, SD-XL~\cite{podell2023sdxl}, or consistency models~\cite{song2023consistency}.


\section{More Experiments}

We will provide more quantitative evaluations and visualization samples of our {\em open-ended visual storytelling} in this section. 

\subsection{Analysis on multi-frame conditioning}
We conduct the multi-frame condition experiments on a test subset with 5,400 samples.
As shown in Table~\ref{tab:multi_frame}, we find that: 
(i) Conditioning on previous frame is critical,
(ii) the number of conditioning frames gives similar results.
However, we do find differences in qualitative results, so we use the 3 closest frames as conditions in auto-regressive generation.
As for frames with less than 3 previous frames, we use all previous frames instead.

\begin{table}[htp]
\scriptsize
\begin{center}
\vspace{-0.2cm}
\begin{tabular}{c|c|c|c|c}
\toprule
& 0-frame & 1-frame & 2-frame & 3-frame \\ 
\midrule
FID $\downarrow$ & 40.29 & \textbf{32.34} & 33.27  & 33.65 \\ 
CLIP-I $\uparrow$ & 0.6841 & 0.7368 & 0.7419 & \textbf{0.7435} \\ 
\bottomrule
\end{tabular}
\end{center}
\vspace{-0.5cm}
\caption{\textbf{Comparison of multi-frame conditioning.}}
\label{tab:multi_frame}
\vspace{-0.5cm}
\end{table}

\subsection{Multi-object conditioned Story Continuation}
As illustrated in Figure~\ref{fig:multi_condition_generation}, 
benefiting from our diverse StorySalon dataset and training strategy, our StoryGen model also demonstrates excellent performance on multi-object story continuation.

\subsection{Story Generation Visualization}

\begin{figure*}[htb]
  \centering
  \includegraphics[height=\textheight]{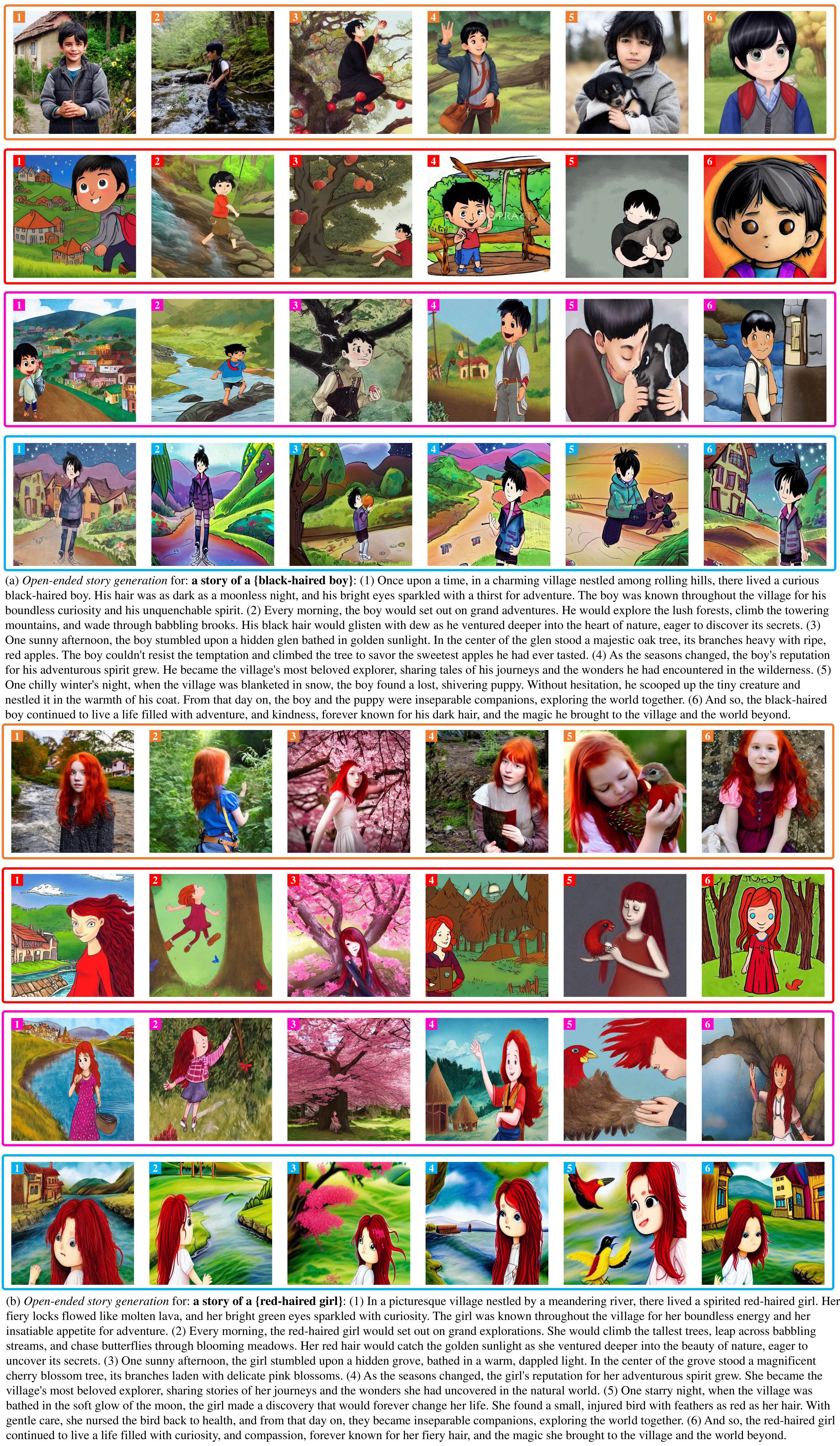} \\
  \vspace{-0.4cm}
  \caption{
  \textbf{Visualization results of Story Generation}.
   The images in \textcolor{orange}{orange}, \textcolor{red}{red}, \textcolor{magenta}{pink}, and  \textcolor{cyan}{blue} boxes are generated by \textcolor{orange}{SDM}, \textcolor{red}{Prompt-SDM}, \textcolor{magenta}{StoryGen-Single}, \textcolor{cyan}{StoryGen}, respectively.
  }
 \label{fig:gen1}
\end{figure*} 

\begin{figure*}[htb]
  \centering
  \includegraphics[height=\textheight]{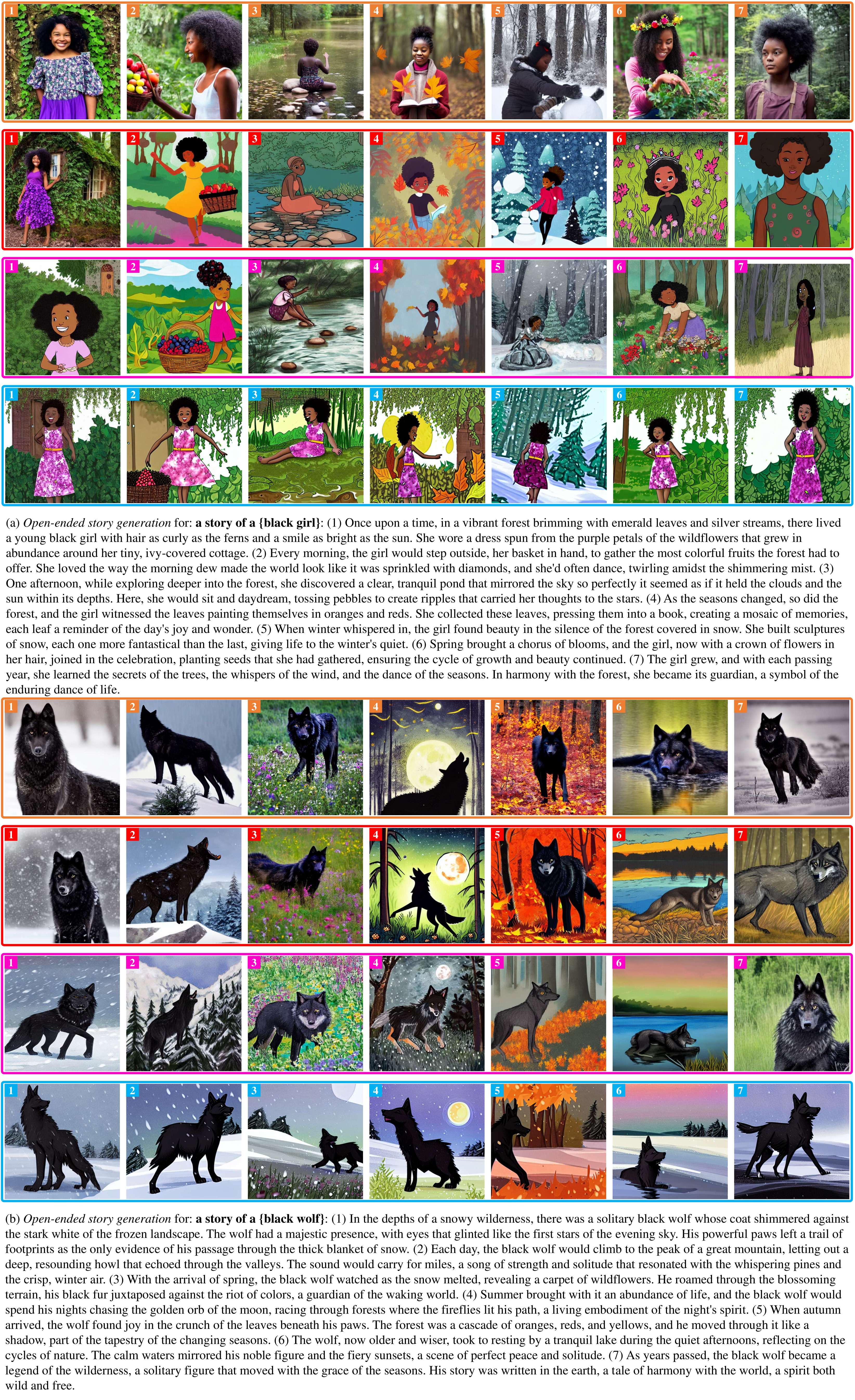} \\
  \vspace{-0.4cm}
  \caption{
  \textbf{Visualization results of Story Generation}.
   The images in \textcolor{orange}{orange}, \textcolor{red}{red}, \textcolor{magenta}{pink}, and  \textcolor{cyan}{blue} boxes are generated by \textcolor{orange}{SDM}, \textcolor{red}{Prompt-SDM}, \textcolor{magenta}{StoryGen-Single}, \textcolor{cyan}{StoryGen}, respectively.
  }
 \label{fig:gen2}
\end{figure*} 

\begin{figure*}[htb]
  \centering
  \includegraphics[height=\textheight]{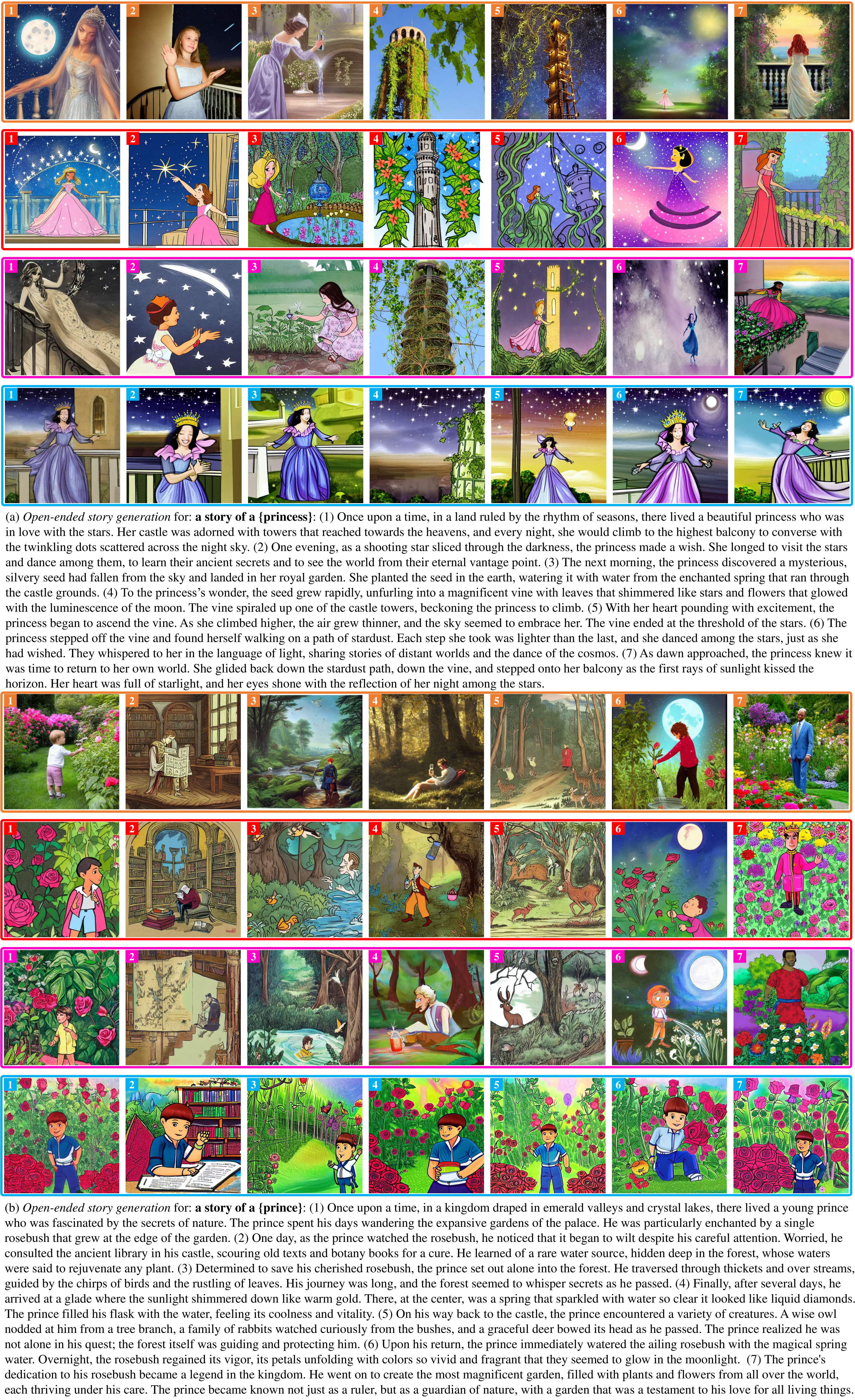} \\
  \vspace{-0.3cm}
  \caption{
  \textbf{Visualization results of Story Generation}.
   The images in \textcolor{orange}{orange}, \textcolor{red}{red}, \textcolor{purple}{purple}, and  \textcolor{cyan}{blue} boxes are generated by \textcolor{orange}{SDM}, \textcolor{red}{Prompt-SDM}, \textcolor{purple}{StoryGen-Single}, \textcolor{cyan}{StoryGen}, respectively.
  }
 \label{fig:gen3}
\end{figure*} 

We conduct a comparative analysis of results from StoryGen against those generated by SDM~\cite{SDM}, Prompt-SDM~\cite{SDM}, and StoryGen-Single.
As illustrated in Figure~\ref{fig:gen1}, Figure~\ref{fig:gen2}, and Figure~\ref{fig:gen3}, the story generation results of our proposed models demonstrate significantly better consistency in style and character, as well as improved alignment between text and image.

While the results from SDM and Prompt-SDM are visually appealing, they exhibit a lack of stylistic and character consistency. The results of StoryGen-Single also display inconsistency, which proves that our StoryGen effectively utilizes the additional image condition to achieve consistency, rather than naive memorization.

\subsection{Story Continuation Visualization}

\begin{figure*}[htb]
  \centering
  \includegraphics[width=\textwidth]{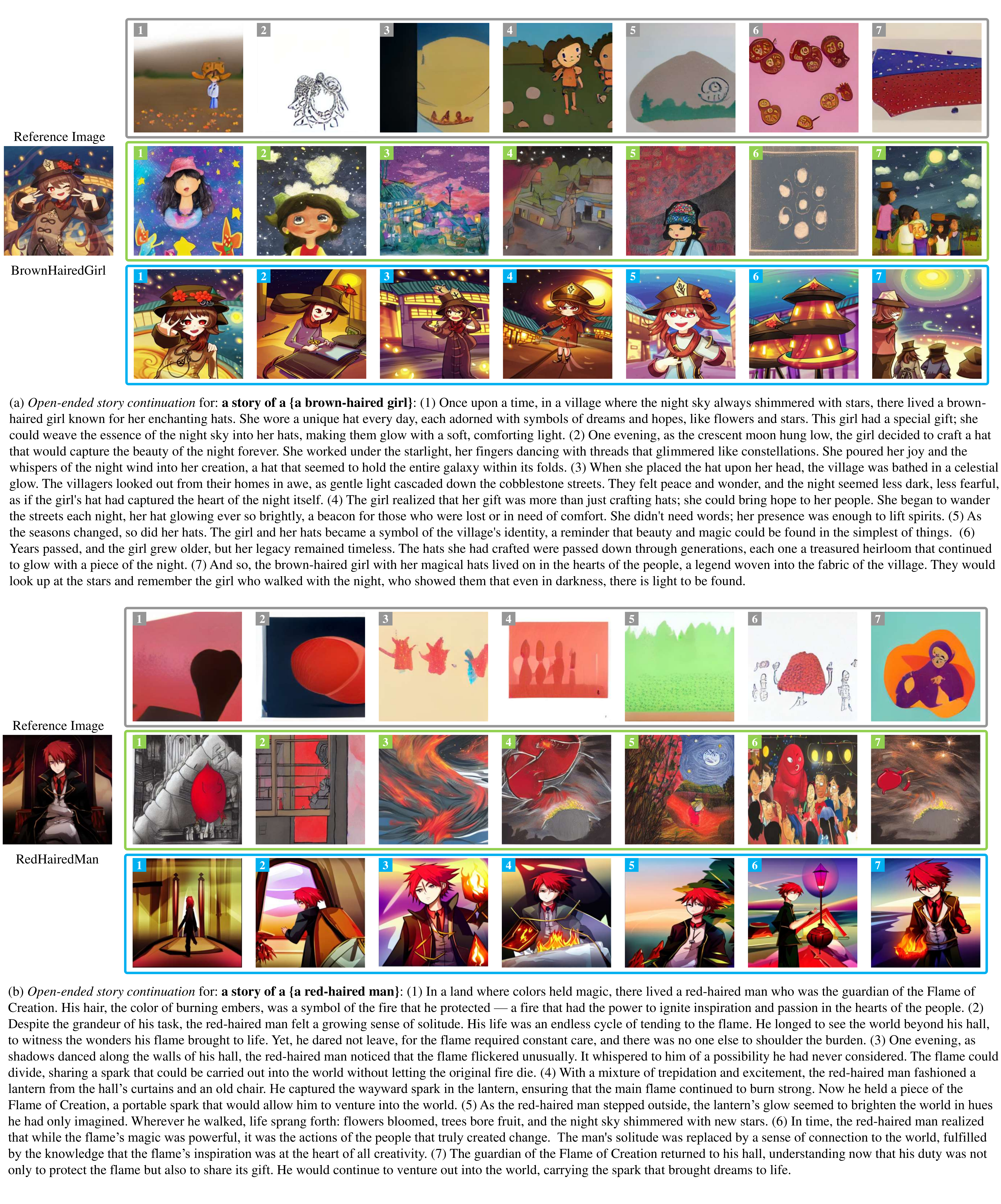} \\
  \vspace{-0.3cm}
  \caption{
  \textbf{Visualization results of Story Continuation}.
   The images in \textcolor{gray}{gray}, \textcolor{lime}{green}, and \textcolor{cyan}{blue} boxes are generated by \textcolor{gray}{StoryDALL-E},  \textcolor{lime}{AR-LDM}, \textcolor{cyan}{StoryGen}, respectively.
  }
 \label{fig:cont1}
\end{figure*} 

\begin{figure*}[htb]
  \centering
  \includegraphics[width=\textwidth]{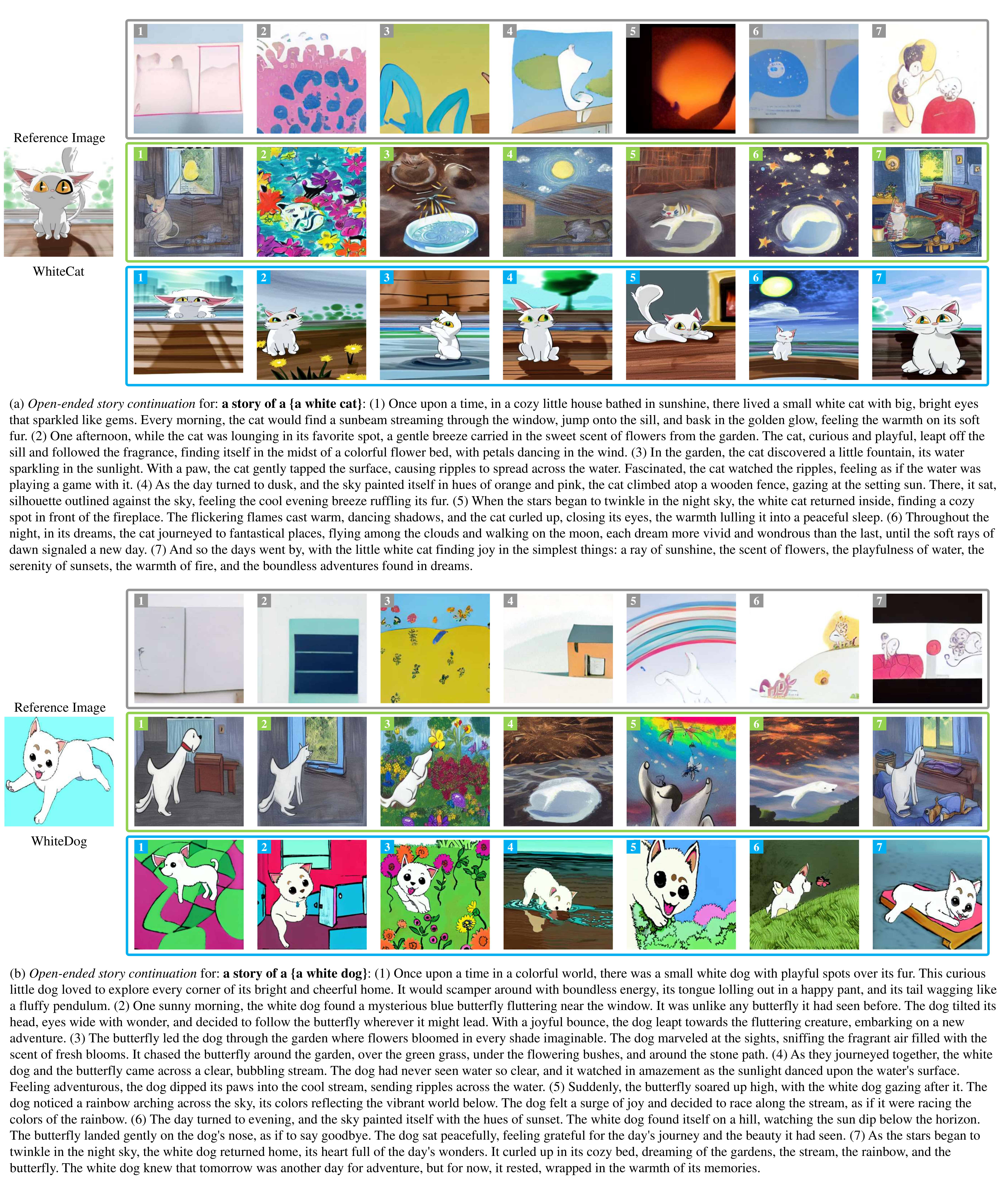} \\
  \vspace{-0.3cm}
  \caption{
  \textbf{Visualization results of Story Continuation}.
   The images in \textcolor{gray}{gray}, \textcolor{lime}{green}, and \textcolor{cyan}{blue} boxes are generated by \textcolor{gray}{StoryDALL-E},  \textcolor{lime}{AR-LDM}, \textcolor{cyan}{StoryGen}, respectively.
  }
 \label{fig:cont2}
\end{figure*} 

\begin{figure*}[htb]
  \centering
  \includegraphics[width=\textwidth]{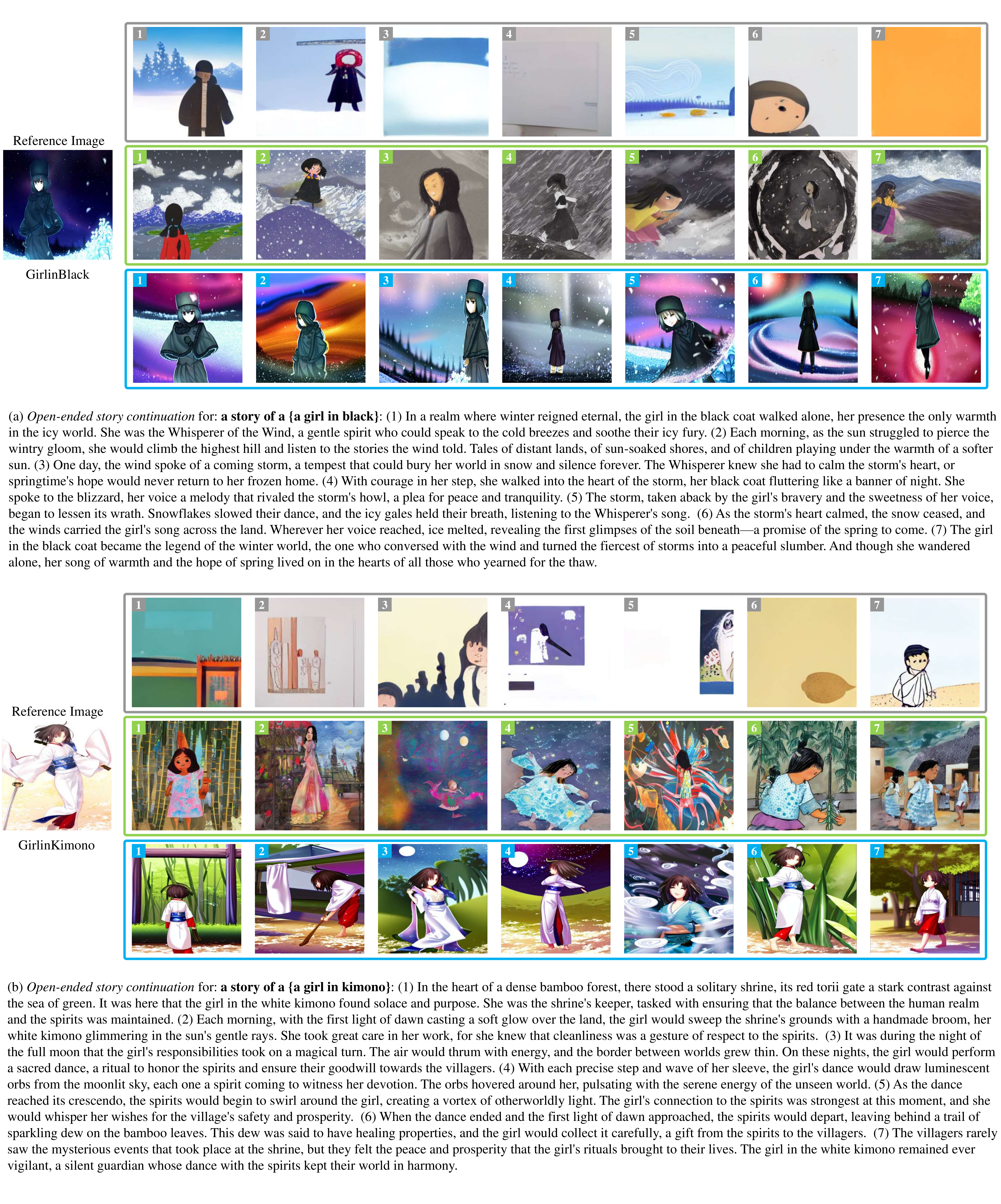} \\
  \vspace{-0.3cm}
  \caption{
  \textbf{Visualization results of Story Continuation}.
   The images in \textcolor{gray}{gray}, \textcolor{lime}{green}, and \textcolor{cyan}{blue} boxes are generated by \textcolor{gray}{StoryDALL-E},  \textcolor{lime}{AR-LDM}, \textcolor{cyan}{StoryGen}, respectively.
  }
 \label{fig:cont3}
\end{figure*} 

We undertake another comparative analysis between the results from StoryGen and those from StoryDALL-E~\cite{maharana2022storydall} and AR-LDM~\cite{pan2022synthesizing}.
As depicted in Figure~\ref{fig:cont1}, Figure~\ref{fig:cont2}, and Figure~\ref{fig:cont3}, the story continuation results of our proposed models exhibit superior proficiency in maintaining style and character consistency, 
achieving stronger alignment between story and image, and enhancing image quality.
Note that, all characters in the given reference image are unseen in our StorySalon datasets.



\clearpage
\subsection{Failure Case Visualization}

\begin{figure*}[!htb]
  \centering
  \includegraphics[width=\textwidth]{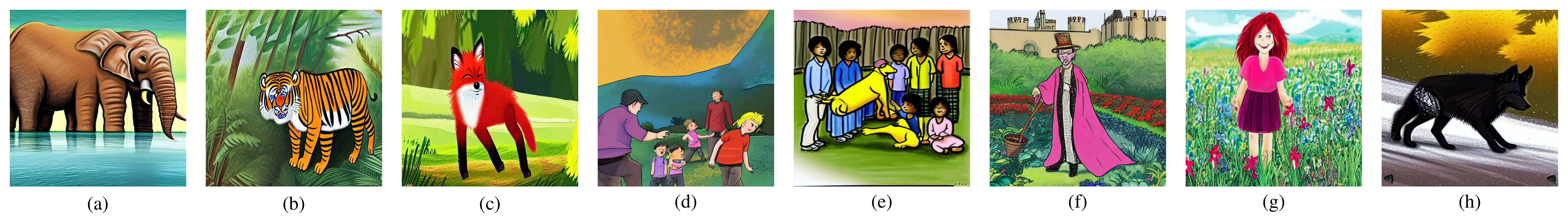} \\
  \vspace{-0.3cm}
  \caption{
  \textbf{Some failure cases of StoryGen.}
  }
 \label{fig:bad}
 \vspace{-6pt}
\end{figure*}

Figure~\ref{fig:bad} presents some instances where StoryGen did not perform optimally.
These failure cases primarily stem from the inherent limitations of SDM.
Figures (a), (b), and (c) illustrate occurrences where StoryGen is prone to generating images with limb count inaccuracies, such as incorrect numbers of legs.
Figures (d) and (e) show scenarios where the generation of multiple objects results in each object being of subpar quality.
Figures (e), (f), and (g) depict instances of StoryGen producing low-quality human faces.
Regarding Figure (h), despite the visual prompt being "{\em A black wolf walking through a forest with autumn leaves falling}", the generated image erroneously includes snowfall, due to the winter setting of the reference image.
This discrepancy arises from the conflict between the image and text conditions.

\end{document}